\documentclass[10pt,journal,compsoc]{IEEEtran}
\ifCLASSOPTIONcompsoc
  \usepackage[nocompress]{cite}
\else
  \usepackage{cite}
\fi
\ifCLASSINFOpdf
\else
\fi
\usepackage{ragged2e}
\usepackage[noend]{algpseudocode}
\usepackage{algorithmicx,algorithm}
\usepackage{cite}
\usepackage{amsmath,amssymb,amsfonts}
\usepackage{graphicx}
\usepackage{textcomp}
\usepackage{xcolor}
\usepackage{CJK}
\usepackage{indentfirst}
\usepackage{flushend}
\renewcommand{\raggedright}{\leftskip=0pt \rightskip=0pt plus 0cm} 
\def\BibTeX{{\rm B\kern-.05em{\sc i\kern-.025em b}\kern-.08em
		T\kern-.1667em\lower.7ex\hbox{E}\kern-.125emX}}

\begin{document}
	
\title{Accelerating Generative Neural Networks on Unmodified Deep Learning Processors - A Software Approach}
\author{Dawen~Xu, ~\IEEEmembership{Member,~IEEE, }Ying~Wang, ~\IEEEmembership{Member,~IEEE, } Kaijie~Tu, ~\IEEEmembership{Member,~IEEE, }Cheng~Liu, ~\IEEEmembership{Member,~IEEE, }Bingsheng~He, ~\IEEEmembership{Member,~IEEE, }Lei~Zhang,~\IEEEmembership{Member,~IEEE}
\IEEEcompsocitemizethanks{
{\IEEEcompsocthanksitem Dawen Xu is with the School of Electronic Science $\&$ Applied Physics, Hefei University of Technology, Anhui, China, 230009.\protect\\
 E-mail: xdw\underline{ }3036@qq.com}
{\IEEEcompsocthanksitem Ying Wang is with the State Key Laboratory of Computer Architecture, Institute of Computing Technology, Chinese Academy of Sciences, Beijing, China, 100089.\protect\\
E-mail: wangying2009@ict.ac.cn}
{\IEEEcompsocthanksitem Kaijie Tu, Cheng Liu and Lei Zhang are with the Research Center for Ubiquitous Computing Systems, Institute of Computing Technology, Chinese Academy of Sciences, Beijing, China, 100089.\protect\\
	E-mail: $\{$tukaijie, liucheng, zlei$\}$@ict.ac.cn}
{\IEEEcompsocthanksitem Bingsheng He is with the Department of Computer Science, School of Computing, National University of Singapore, Singapore, 119260.\protect\\
	E-mail: hebs@comp.nus.edu.sg}
}

\thanks{Manuscript received December 15, 2019.}
}

\IEEEtitleabstractindextext{%
	
\begin{abstract}
\raggedright Generative neural network is a new category of neural networks and it has been widely utilized in applications such as content generation, unsupervised learning, segmentation and pose estimation. It typically involves massive computing-intensive deconvolution operations that cannot be fitted to conventional neural network processors directly. However, prior works mainly investigated specialized hardware architectures through intensive hardware modifications to the existing deep learning processors to accelerate deconvolution together with the convolution. In contrast, this work proposes a novel deconvolution implementation with a software approach and enables fast and efficient deconvolution execution on the legacy deep learning processors. Our proposed method reorganizes the computation of deconvolution and allows the deep learning processors to treat it as the standard convolution by splitting the original deconvolution filters into multiple small filters. Compared to prior acceleration schemes, the implemented acceleration scheme achieves 2.41$\times$ - 4.34$\times$ performance speedup and reduces the energy consumption by 27.7\% - 54.5\% on a set of realistic benchmarks. In addition, we also applied the deconvolution computing approach to the off-the-shelf commodity deep learning processors. The performance of deconvolution also exhibits significant performance speedup over prior deconvolution implementations. 
\end{abstract}

\begin{IEEEkeywords}
Generative neural network, deconvolution accelerator, split deconvolution.
\end{IEEEkeywords}}

\maketitle
\IEEEdisplaynontitleabstractindextext
\IEEEpeerreviewmaketitle

\section{Introduction}
\IEEEPARstart{D}eep neural networks are making continuous breakthroughs in massive research territories over the years. In contrast to the conventional convolutional neural networks heavily utilized for object classification and detection, generative neural networks \cite{goodfellow2014generative} have been proved to be superior in a broad domain of applications including content-generation, unsupervised learning, segmentation and pose estimation. Typically, the generative neural networks involve both convolutional layers and deconvolutional layers. Both layers are compute-intensive and are the performance bottleneck of generative neural networks. Therefore, it is demanded to accelerate the backbone architecture of the networks, especially the generative networks on end-devices for real-time and low power applications such as real-time deepfake \cite{korshunova2017fast} and style transfer \cite{lengstrom2016faststyletransfer}. For exemplary generative neural network benchmarks described in Table \ref{tab1}, the deconvolution layers contribute to the major overhead of the multiply-and-add operations in the benchmark (The total operands refer to those of the inference phase). The deconvolution operation is used as an indispensable component to restore the condensed feature maps to full-size at the top of the networks, which are the common architectures in generative networks and other popular models used for semantic segmentation and instance detection \cite{godard2017unsupervised}.

\par
\begin{figure}[b]
	\centering{\includegraphics[width=.94\columnwidth]{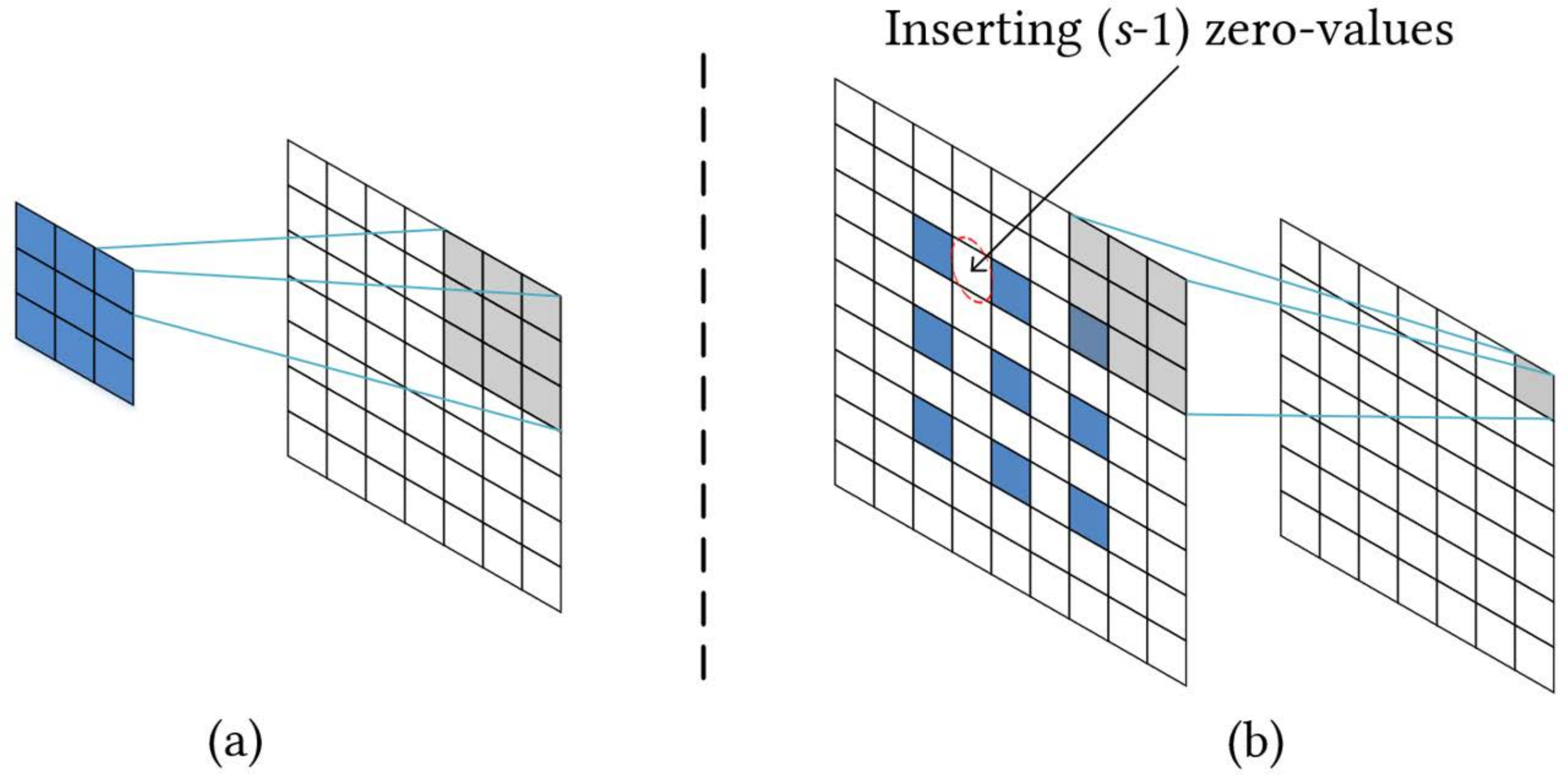}}
	\caption{Computational process of (a) deconvolution and (b) deconvolution with inserted zero-values.}
	\label{fig1}
\end{figure}

\setlength\parindent{1em}Hardware specialization is a popular approach to accelerate the computation of neural network based applications. To accelerate generative neural networks with customized hardware other than general purpose compute units, researchers have tried a number of approaches from distinct angles. For more efficient design, an intuitive solution is to reuse the convolution processor and build a unified fully convolutional processor for both convolution and deconvolution operations. In such architectures input data of deconvolution can be reorganized by dynamically padding zero activations to the original feature maps and then treat the deconvolution as the conventional convolution layer as presented in Figure \ref{fig1}. Figure \ref{fig1}(a) is an example of the classic deconvolutional operation with the stride of 2, while Figure \ref{fig1}(b) is converted equivalent convolutional operation with stride set to be 1. Eventually, the deconvolution can be mapped to the convolution processor without any hardware modification. However, the zero activations induce considerable redundant computing and degrade the performance which is illustrated in \cite{xu2018fcn}. Although many CNN accelerators \cite{chen2016eyeriss,chen2014diannao,xu2018fcn,chen2014dadiannao,song2016c,albericio2016cnvlutin}are able to skip the zero activations during the computing through additional zero detection logic, they typically can only skip a portion of the zero activations especially the ones that are located on the boundary of the feature maps. However, the zero padding deconvolution approach as shown in Figure \ref{fig1}(b) has many zero activations inserted between the non-zero activations and they are usually difficult to be removed due to aligned computing data flow on the parallel computing units in DNN accelerators.

To improve the computing efficiency of deconvolution, the authors in \cite{zhang2017design} opted to build independent processor engines for convolution and deconvolution operation respectively. This approach raises a large portion of hardware resources and chip area increase.  Different from the above two approaches, the authors in \cite{xu2018fcn} and \cite{yazdanbakhsh2018ganax} proposed to revisit the convolutional processor and change the micro-architecture to support both convolution and deconvolution efficiently in a unified processor. In addition, these methods also need dedicated data flow scheduler to make use of the computing engine. For unified architectures, the advantage is better performance and hardware utility, while the disadvantage is the additional redesign and engineering cost. However, for off-the-shelf CNN processors without specialized deconvolution support such as Diannao \cite{chen2014diannao} and TPU \cite{jouppi2017datacenter}, the inefficiency and resource under-utilization induced by the zero-padding approach is an inevitable cost to implement the deconvolutional layers.
\begin{table}[t]
	\caption{The number of multiply-add operations in the inference phase.}
	\centering{\includegraphics[width=.94\columnwidth]{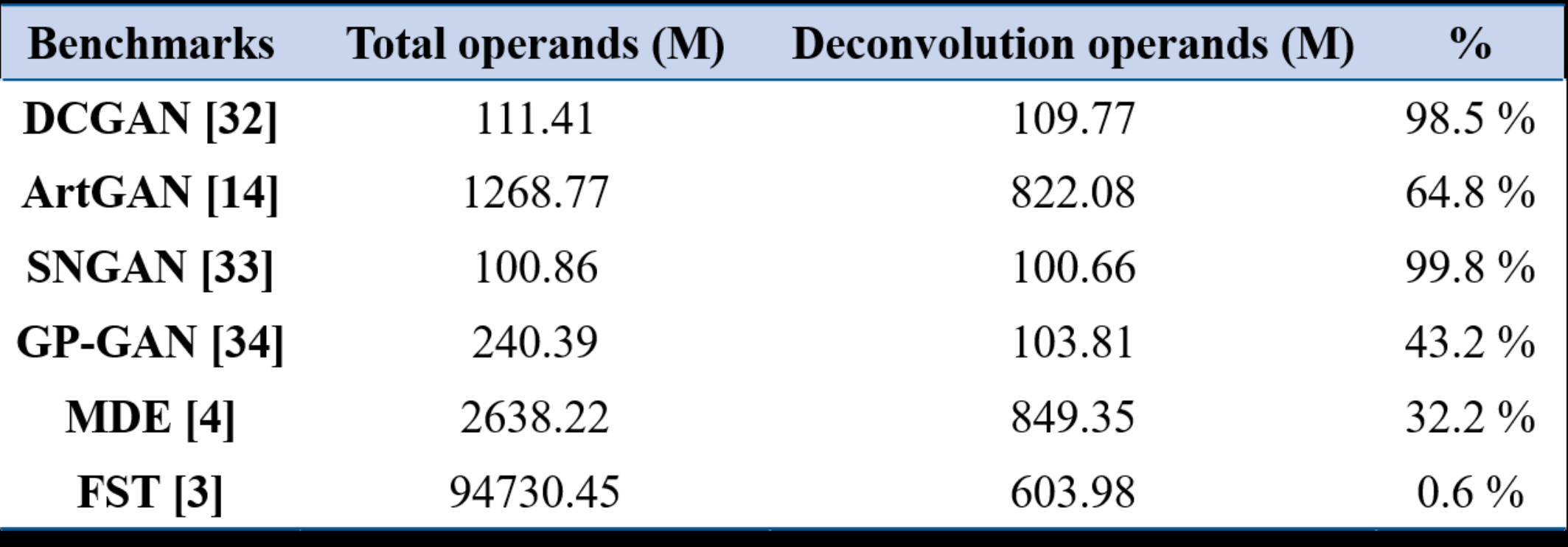}}
	\label{tab1}
\end{table}
\par\setlength\parindent{1em}Inspired by the prior work, we seek to support fast and efficient deconvolution layer implementation on general CNN processors like Eyeriss \cite{chen2016eyeriss}, Diannao \cite{chen2014diannao} and TPU \cite{jouppi2017datacenter}, some of which are already commercialized and widely-used in different areas. For these classic CNN processors, many zero-value activations must be padded to the feature map in order to map the deconvolution layers on to them. Instead of zero-padding that induces numerous redundant computing operations, we tailor a novel implementation of deconvolution layer from the software angle, and pre-partition the deconvolutional filters into multiple small convolutional filters, so that the deconvolution operations are converted and can be efficiently implemented on any CNN processor without redesigning or replacing them. In our evaluation on classic CNN processors, the performance as well as the energy efficiency of our deconvolution implementation remains competitive compared to prior work of specialized GAN processors.

\par\setlength\parindent{1em}In summary, our contributions can be summarized as follows:
\begin{itemize}
	\item We proposed a novel filter partitioning and reorganization approach to convert a general deconvolution operation to multiple standard convolution operations strictly without incurring much computing redundancy such that deconvolution can be implemented efficiently as convolution.
	
	\item We investigated the way to reorganize the split deconvolution results efficiently on legacy neural network processors without hardware modification.
	
	\item We evaluated the proposed deconvolution performance on a set of representative benchmarking networks with comprehensive experiments, the experiments show that the proposed approach achieves competitive performance over the state-of-the-art deconvolution processors on both general CNN processors and the most advanced commodity deep learning processor chips such as Google TPU and Intel Neural Compute Stick 2, which are released recently.
\end{itemize}

\par\setlength\parindent{1em} The rest of this paper is organized as follows. Section II presents the related work of deconvolution acceleration design. Section III describes the architecture of typical CNN processors. In Section IV, we elaborate the conversion process of generic split deconvolution in detail. At length, Section V presents the evaluation results and Section VI concludes the paper.
\section{Related work}
\par\setlength\parindent{1em}With the advancements of deep learning, various neural networks have been proposed to address different tasks such as objection detection and image classification. Among them, generative neural networks are demonstrated to be particularly efficient for content generation tasks like image style transfer \cite{tan2017artgan, godard2017unsupervised, lengstrom2016faststyletransfer}, segmentation tasks such as \cite{chen2014semantic}, and pose estimation tasks such as \cite{newell2016stacked}. These novel neural networks attract a lot of attentions. Ledig et al. \cite{ledig2017photo} proposed SRGAN and adopted a perceptual similarity loss to generate detailed images from low-resolution images. By using generative adversarial networks (GANs), high-resolution images of small objects can be generated and utilized to improve target detection accuracy \cite{li2017perceptual}. Generative neural networks can also be applied for sequence data generation as presented in SeqGAN \cite{yu2017seqgan} and ORGAN \cite{guimaraes2017objective}. Additionally, more variants of generative neural networks have been developed and employed in semi-supervised learning and the medical field \cite{chongxuan2017triple,yang2017automatic}.

\par\setlength\parindent{1em}However, generative neural networks that consist of both compute-intensive convolution and deconvolution operators cannot be fitted to the conventional CNN processors directly \cite{chen2016eyeriss,chen2014diannao,chen2014dadiannao,wang2016deepburning}. As deconvolution is also computing intensive and hinders the acceleration of generative neural networks on CNN processors, thereby, it is highly demanded to explore hardware acceleration of deconvolution operations. Zhang X et al. in \cite{zhang2017design} proposed to optimize deconvolution with reverse looping and stride hole skipping. Despite the excellent performance, combining independent convolution and deconvolution components in an processor induces considerable chip area and power consumption. Amir Y et al. in \cite{yazdanbakhsh2018ganax} proposed to convert deconvolution to convolution by adding zero padding to the activations and then developed a unified MIMD-SIMD processor for both operations. In addition, it implemented a set of distributed on-chip buffers to avoid the redundant computing brought by the inserted zero activations. Based on \cite{yazdanbakhsh2018ganax}, the authors further developed an end-to-end template-based solution, which can generate the optimized synthesizable unified processor from a high-level specification of GANs in \cite{yazdanbakhsh2018flexigan}. Instead of adding zeros to input feature map, Xu et al. in \cite{xu2018fcn} proposed a unified FCN processor on top of a bi-direction systolic array. The FCN processor performs the computing on original input features. The weight and data of adjacent PEs are shared and passed periodically by taking advantage of the small column buffers added to the 2D PE array. Similar to \cite{xu2018fcn}, Wang et al. in \cite{wang2019towards} designed a uniform architecture to support both 2D and 3D deconvolutional neural networks on FPGAs. Multiple FIFOs are added to adjacent PEs to deliver the overlapped temporary results. Yan et al. in \cite{yan2018gna} proposed a cold buffer to store the overlapped computing results for more efficient data reuse and a novel mapping approach to improve the utilization of the computing array for both convolution and deconvolution. Intel NCS2 \cite{ncs2} also has specialized hardware to support native deconvolution, but there are no open technical details. In summary, it can be found that hardware redesigning is typically required to have existing CNN processor to support deconvolution computing in generative neural networks. Due to the long hardware design cycle, many commodity neural network processor including Google Edge TPU \cite{jouppi2017datacenter} and Ropal Neural Compute Stick Lightspeeur SPR2801 \cite{ropal} still do not support the raw generative neural networks yet. Besides the mainstream solutions that accelerating neural networks on ASIC or FPGAs, the implementations on Resistive Random Access Memory (ReRAM) are also one of the recent research hotpots. F. Chen et al. \cite{8806783} accelerates GANs using a filter deformation method to completely eliminate inserted zeros in deconvolutional layers based on 3D horizontal ReRAM architecture. However, due to the technological limitations, the implementation of neural networks accelerators on ReRAM is still being explored. The purpose of this paper is to propose a acceleration method for GANs to be reused on existing mainstream accelerators.
\begin{figure}[t]
	\centerline{\includegraphics[height=6cm,width=7.9cm]{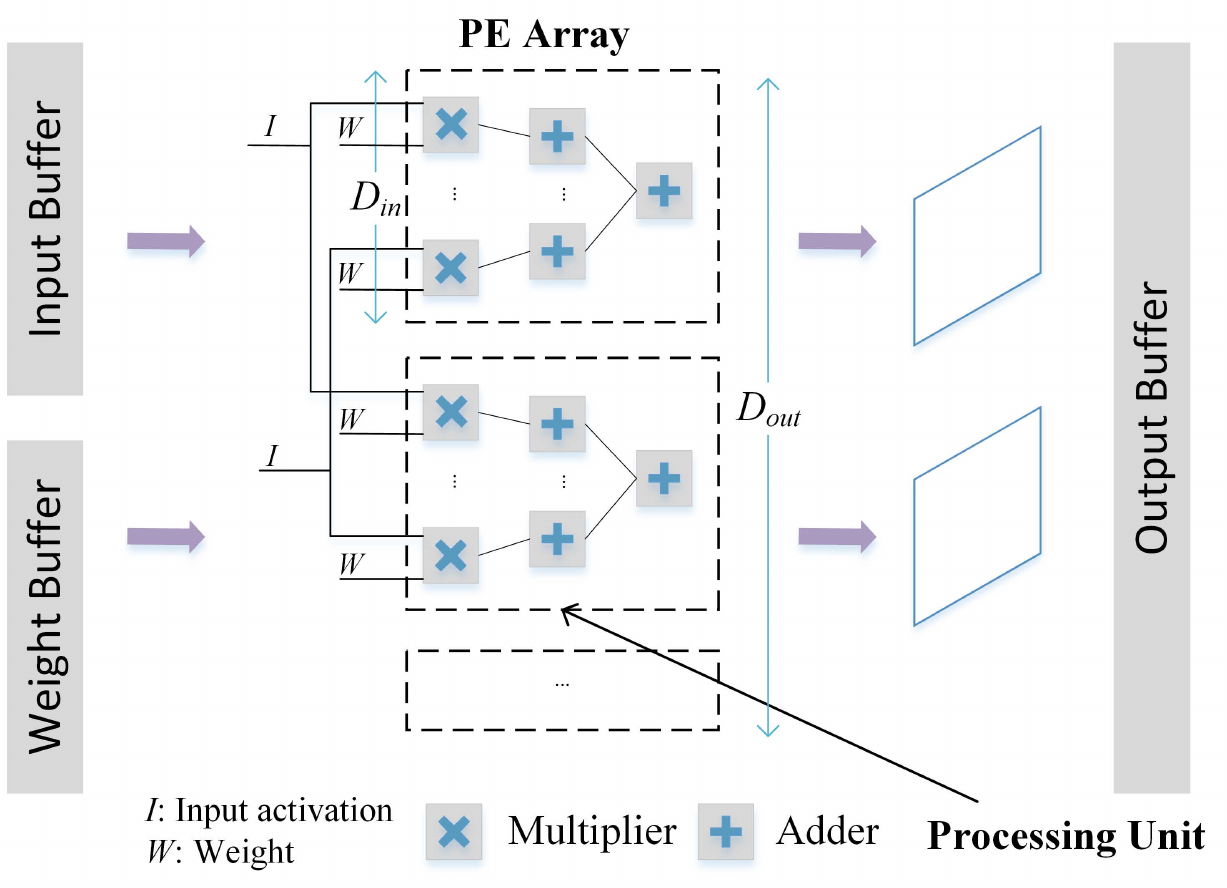}}
	\caption{Dot-production based CNN processor \cite{chen2014diannao,chen2014dadiannao,song2016c,albericio2016cnvlutin}}
	\label{fig2}
\end{figure}

\par\setlength\parindent{1em}Different from the above works, researchers seek to reuse the conventional CNN processors for generative neural networks without hardware redesigning. Shi et al. \cite{shi2016deconvolution} presented a simple example of transformation from deconvolution to convolution by padding zeros to the input feature maps. However, the fixed zero-padding to the right and bottom of the input features only works for the first partition of the split deconvolution and it can cause errors when this zero-padding is utilized for the deconvolution conversion. The correct padding must be adapted to the deconvolution partition as well as the output feature cropping strategies to ensure equivalent output to the raw deconvolution. In addition, this work is posted as a blog with limited experiments and has not gone through the peer review. Chang et al. \cite{chang2018optimizing} utilized filter deformation and proposed an approximate conversion approach targeting at super-resolution image reconstruction problems. While super-resolution image reconstruction can typically tolerate computing errors, the approximate conversion approach works fine but it cannot be applied to general generative neural networks that are not necessarily fault-tolerant. In addition, the approach proposed in \cite{chang2018optimizing} needs to rearrange the deconvolutional results on CPU instead of CNN processors and it can cause massive data communication between CPU and the processors. To address the above problems, we aim to develop a software approach that can deploy generative neural networks directly on the existing CNN processors without precision penalty nor hardware modification.
\begin{figure}[t]
	\centerline{\includegraphics[height=4.8cm,width=8.8cm]{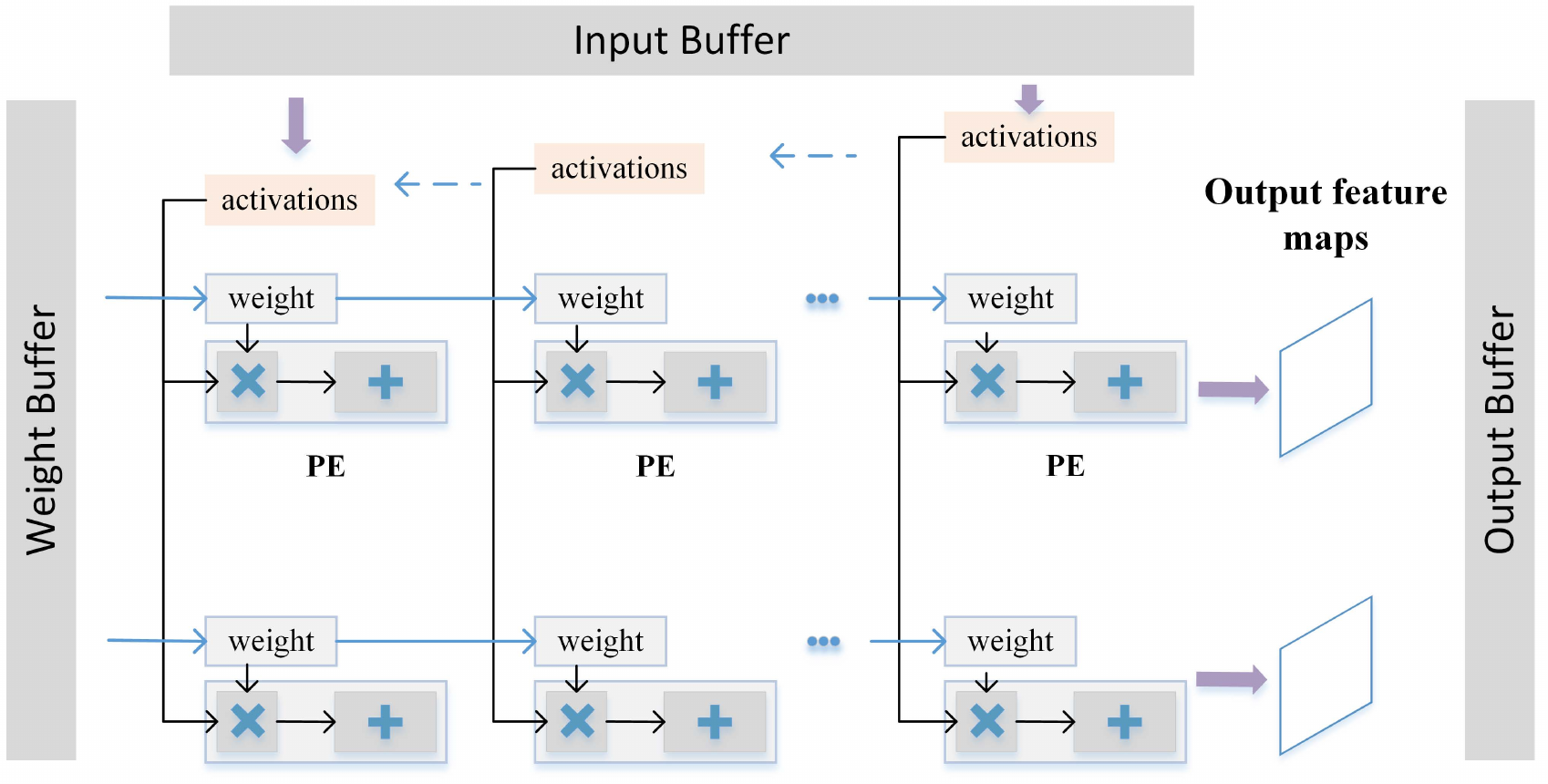}}
	\caption{Regular 2D array CNN processor \cite{chen2016eyeriss,xu2018fcn,jouppi2017datacenter}}
	\label{fig3}
\end{figure} 
\section{Typical CNN Processors}
\par\setlength\parindent{1em}This section briefly explains the architecture of the two  mainstream architectures of general CNN processors assumed in this paper, including dot-production array processor and regular 2D array processor. Most of the prior CNN processors can be included in these two typical architectures \cite{chen2016eyeriss,chen2014diannao,jouppi2017datacenter,xu2018fcn,chen2014dadiannao,song2016c,albericio2016cnvlutin}.

\subsection{Dot-production array processor}
\par\setlength\parindent{1em}Figure \ref{fig2} shows the dot-production based CNN processor. It consists of \textit{D$_{out}$} neural processing units.Each neural process unit includes \textit{D$_{in}$} multipliers as well as an adder tree and performs a dot production.The same \textit{D$_{in}$} input activations are fed concurrently to each processing unit per cycle while the weights are different. Each unit accepts \textit{D$_{in}$} weights per cycle and \textit{D$_{in}$} $\times$ \textit{D$_{out}$} parameters needs to be sent to the array per cycle. In each PE, \textit{D$_{in}$} partial results obtained from multipliers are consumed by the adder tree, and a dot production can be completed each cycle because of the pipelined processing architecture. Once a filter window is processed, the output result is sent to an activation function unit and the result is transferred to the output buffer. Each output activation produced by the processing unit belongs to a different output channel. When weight or data cannot be accommodated by the on-chip buffers, the neural networks will be tiled to fit to the architecture. Diannao \cite{chen2014diannao}, Dadiannao \cite{chen2014dadiannao}, C-brain \cite{song2016c} and Cnvlutin \cite{albericio2016cnvlutin} are typical designs that adopt the dot-production array architecture.

\subsection{Regular 2D array processor}
\par\setlength\parindent{1em}Another typical CNN processor architecture with regular 2D PE array is illustrated in Figure \ref{fig3}. Compared to the former structure, it mainly differs on the data flow. The data flow used in this work is output stationary (OS) according to the definition in Eyeriss \cite{chen2016eyeriss}. Basically, each PE in the array performs all the operations required to yield an output activation. The weights are fed from the first column of the array and flow across the PEs from left to right to guarantee that all PEs operate in full scale. The input activations are broadcast to all the PEs in a column, but we have at most one PE column to receive the input activations alleviating the pressure on on-chip buffers bandwidth. Each row of the PE array produces output activations of one output feature map on y-axis. Each column PE produces the output activations belonging to different output feature maps but the same pixel positions. Under the circumstances, both input activations and weights consume a limited amount of on-chip memory bandwidth. This architecture enables the proposed system to achieve high reusability and eventually benefits more on boosting throughput in a limited bandwidth provision. This feature is handy in reducing the bandwidth demand caused by weight/activation load and store, especially when the processors are sharing the on-chip storage space and bandwidth with other application processors in nowadays heterogeneous SoCs adopted in mobile and embedded systems. Eyeriss \cite{chen2016eyeriss}, TPU \cite{jouppi2017datacenter},  FCN-Engine \cite{xu2018fcn} are typical designs that adopt the 2D array architecture

\begin{figure}[h]
	\centerline{\includegraphics[height=9.0cm,width=8.8cm]{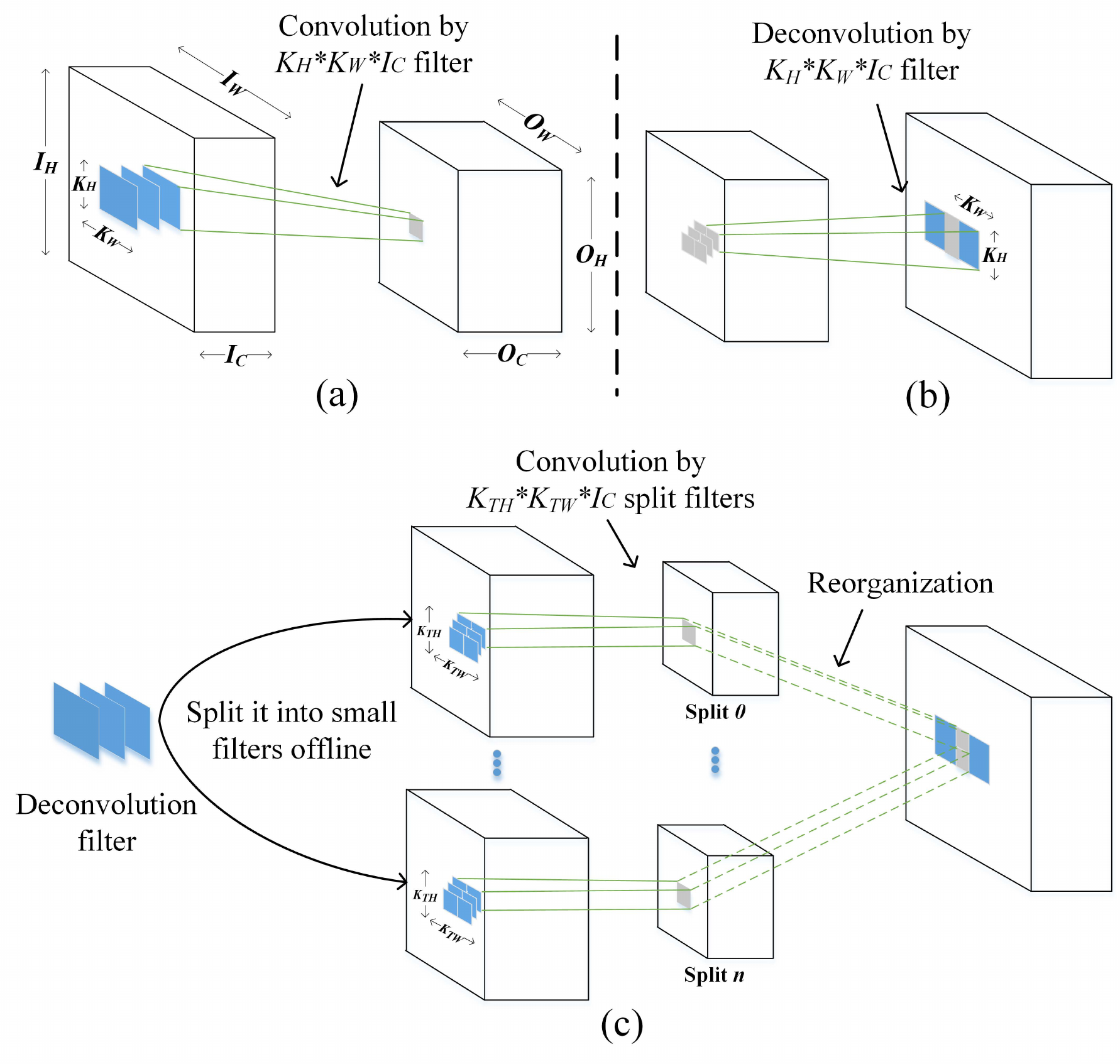}}
	\caption{ (a) Convolutional layer (b) Deconvolutional layer (c) Split deconvolution that converts a deconvolution layer to multiple convolution layers}
	\label{fig4}
\end{figure}

\begin{algorithm}[bh]
	\caption{Convolution \& Deconvolution}
	\begin{algorithmic}[1] 
		\Require $o_h, o_w, o_c, I_C, K_H, K_W$
		\Ensure $output(o_h,o_w,o_c)$
		
		\Function {Conv}{$o_h, o_w, o_c, I_C, K_H, K_W$}
		\For {$(i_c = 0; i_c < I_C; i_c++)$}
		\For {$(k_h = 0; k_h < K_H; k_h++)$}
		\For {$(k_w = 0; k_w < K_W; k_w++)$}
		\State $output(o_h,o_w,o_c)$ += $input(o_h \times s + k_h, o_w \times s + k_w, i_c) \times w(k_h, k_w, i_c, o_c)$
		\EndFor
		\EndFor
		\EndFor
		\EndFunction
		
		\Function {Deconv}{$o_h, o_w, o_c, I_C, K_H, K_W$}
		\State $left~=~max(0,ceil((o_w~-~K_W~/~s)))$
		\State $right~=~min(I_W~-~1,left~+~ceil(K_W~/~s))$
		\State $top~=~max(0,ceil((o_h~-~K_H~/~s)))$
		\State $bottom~=~min(I_H~-~1,top~+~ceil(K_H~/~s))$
		\For {$(i_c = 0; i_c < I_C; i_c++)$}
		\For {$(i_h = top; i_h < bottom; i_h++)$}
		\For {$(i_w = left; i_w < right; i_w++)$}
		\State $output(o_h,o_w,o_c)$ += $input(i_h, i_w, i_c) \times w(o_h-i_h \times s, o_w-i_w \times s, i_c, o_c)$
		\EndFor
		\EndFor
		\EndFor
		\EndFunction
	\end{algorithmic}
\end{algorithm}

\section{The Proposed Split Deconvolution}
\par\setlength\parindent{1em}In Section IV A, we analyze the correlation between the convolution and deconvolution and brief the idea of converting a deconvolution operation to generic convolutions. Then we present the detailed conversion steps from generic deconvolution operations to standard convolution operations in Section IV B.
\subsection{Correlation between Convolution and Deconvolution }
\par\setlength\parindent{1em}Convolution and deconvolution are the major sources of overhead in generative neural networks. Figure \ref{fig4}(a) and \ref{fig4}(b) show the basic computing patterns of the two operations. In convolution i.e. Figure \ref{fig4}(a), windows of input features are convolved with the corresponding filters first. Then the results are added up to obtain an output element of the output feature. In deconvolution i.e. Figure \ref{fig4}(b), each element of the input feature maps is multiplied to each weight matrix first. Then the production in the overlapped position will be accumulated as the final output activation. By definition, convolution and deconvolution is completely different.

\par\setlength\parindent{1em}In order to reuse the conventional CNN processors for deconvolution operations, we further analyze the computing patterns of convolution and deconvolution. The pseudo code of the two operations for computing one output activation are presented in Algorithm 1. Note that \textit{I$_C$} and \textit{O$_C$} indicate the input and output channel of the feature map. \textit{I$_H$} and \textit{I$_W$} denote the length and width of the input feature map, and \textit{O$_H$}, \textit{O$_W$} are the length and width of the output feature map. \textit{K$_H$} and \textit{K$_W$} is the length and width of the filter. \textit{s} refers to stride. The notations will be used through this paper. Basically, convolution can be computed with an elementwise approach, while deconvolution is consist of multiple group convolution. Each output activation of convolution i.e. \textit{output(o$_h$,o$_w$,o$_c$)} is the accumulation of production of input feature windows ([\textit{o$_h$} $\times$ \textit{s}, \textit{o$_h$} $\times$ \textit{s} + \textit{K$_H$} ), [\textit{o$_w$} $\times$ \textit{s}, \textit{o$_w$} $\times$ \textit{s} + \textit{K$_W$}))with consecutive weight matrices. For deconvolution, each output activation is also the accumulation of production of input feature map window and a set of weights using the same computing function $Convolution$ except that the weights are selected with stride \textit{s} and reassigned new coordinates in the \textit{nth} group. Meanwhile, the output belonging to different groups needs to be reorganized in the final output feature map.

\par\setlength\parindent{1em}With this observation, we proposed a split deconvolution approach as shown in Algorithm 2 which divides the deconvolution filters into multiple smaller filters with stride \textit{s}. In this case, the split filters become consecutive and each deconvolution operation is converted to multiple standard convolution operations. Accordingly, deconvolution can be deployed on conventional CNN processors without any hardware modification. While we need to split the filter and reorganize the obtained activations, detailed conversion approach will be elaborated in the next subsection.

\begin{algorithm}[t]
	\caption{Split Deconvolution}
	\begin{algorithmic}[1] 
		\Require $o_h, o_w, o_c, I_C, K_H, K_W$
		\Ensure $output(o_h,o_w,o_c)$
		
		\Function {SplitDeconv}{$o_h, o_w, o_c, I_C, K_H, K_W$}
		\State {// weights of the $N$ identical split }
		\State {// convolution: $w$($K_{TH}, K_{TW}, I_C$)}
		
		\For{$(n = 0; n < N; n++)$}
		\For{$(i_c = 0; i_c < I_C; i_c++)$}
		\State $ k_{th} \gets K_{TH} - 1$
		\For{$(k_{h} = \lfloor n/s \rfloor; k_{h} < K_{H}; k_h+=s)$}
		\State $ k_{tw} \gets K_{TW} - 1$
		\For{$(k_{w} = n\%s; k_{w} < K_{W}; k_w+=s)$}
		\State $w(n,k_{th},k_{tw},i_{c}) \gets w(k_{h},k_{w},i_{c})$
		\State $ k_{tw} \gets k_{tw} - 1$
		\EndFor
		\State $ k_{th} \gets k_{th} - 1$
		\EndFor
		\EndFor
		
		\State $Conv (o{^n}{_h}, o{^n}_w, o{^n}_c, I_C, K_{TH}, K_{TW})$
		\State Reorganize the obtained $nth$ output activation
		\EndFor	
		\EndFunction
	\end{algorithmic}
\end{algorithm}

\begin{figure}[t]
	\centerline{\includegraphics[height=8.2cm,width=8.8cm]{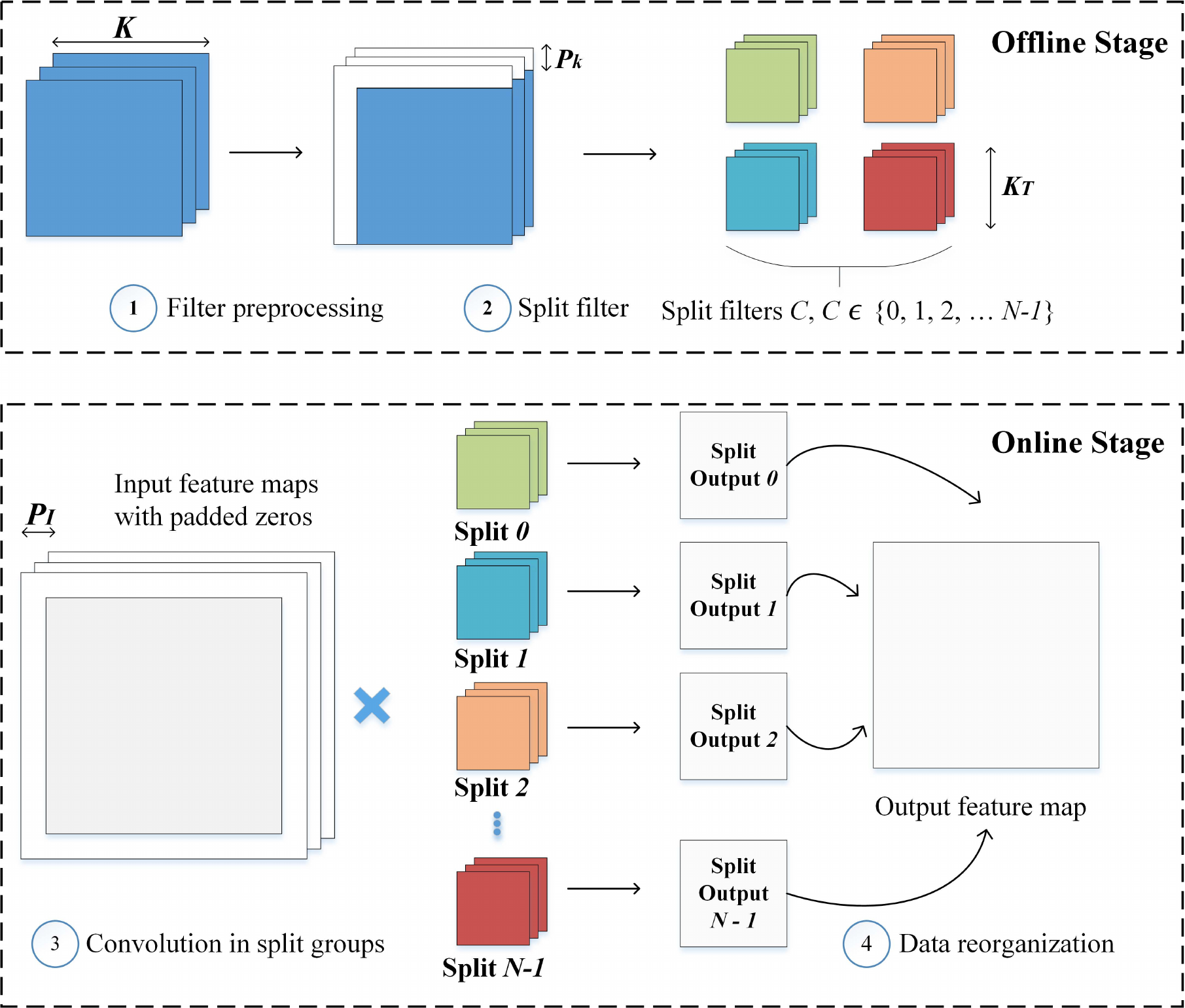}}
	\caption{Conversion steps from deconvolution to convolution, it consists of four steps. 1) The filter is expanded when the filter size is not divisible by the stride. 2) Split the deconvolution filters to multiple small filters according to Equations (\ref{eq4}-\ref{eq6}). 3) The padded input feature maps convolve with the split filters. 4) Reorganize the split deconvolution results to construct the expected deconvolution output by Equations (\ref{eq10}) and (\ref{eq11}). }
	\label{fig5}
\end{figure}

\subsection{Generic Deconvolution Conversion}
\par\setlength\parindent{1em}Following the above idea, we convert generic deconvolution operation to a set of independent convolution operations. The conversion roughly consists of four steps, as shown in Figure \ref{fig5}.
\par\setlength\parindent{1em}The first step is the weight preprocessing in which the original deconvolutional filters will be expanded with zeros on the top and left side when its length and width is not divisible by stride \textit{s}. It ensures that the deconvolution can be converted to multiple identical convolution operations. The padded zeros will expand the output accordingly while the orientation of the padded zeros guarantees that the center of the expanded output covers the standard deconvolution output. The expanded length and width \textit{P$_K$} can be calculated with Equation (\ref{eq1}) where \textit{K$_T$} is the split filter size (assuming it is square) and can be obtained from Equation (\ref{eq2}).

\begin{equation}
P_{K} = s \times K_{T} - K\label{eq1}
\end{equation}

\begin{equation}
K_{T} = ceil(K~/~s)\label{eq2}
\end{equation}

\par
\begin{figure}[t]
	\centerline{\includegraphics[height=7.8cm,width=8.8cm]{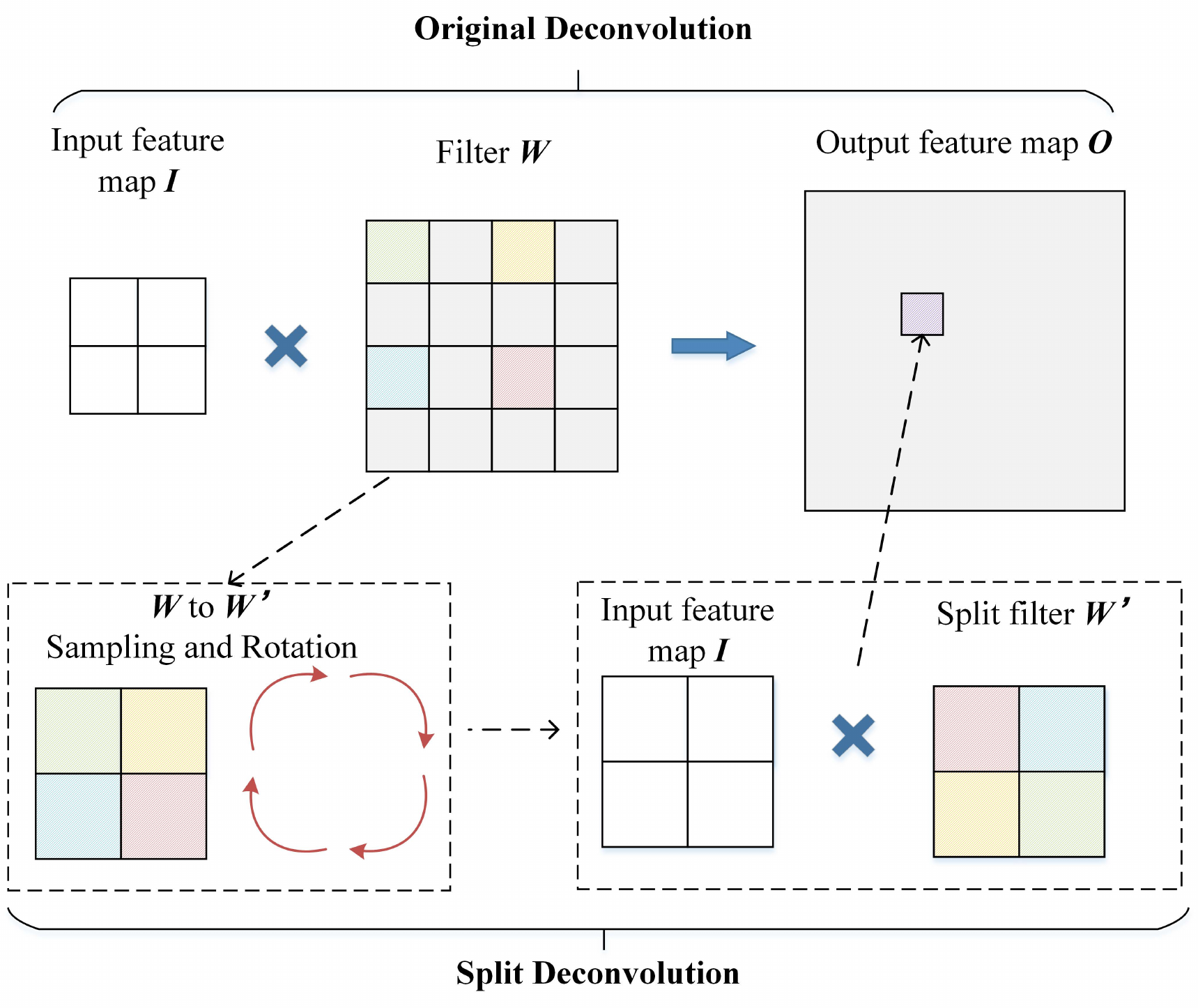}}
	\caption{Weights distribution for an output activation in original deconvolution and split deconvolution where the filter is 4 by 4 and the stride is 2. }
	\label{fig6}
\end{figure}

\setlength\parindent{1em}The second step is to split the deconvolution filters into multiple small filters with sampling and rotation. Figure \ref{fig6} illustrates the coordinate distribution of filters before and after the conversion with a small but representative example.  To compute an output deconvolution activation with standard convolution operations, filters need to be sampled with stride \textit{s} and reorganized into new filters. In addition, each sampled filter needs to be rotated 180 degrees to ensure correct computing. Equation (\ref{eq3}) presents the generic conversion. Each deconvolution will be split into \textit{s$^2$}  convolution operations. The stride of the split convolution operations is constant 1.  Without loss of generality, suppose \textit{W$_n$} is the nth convolutional filter. It can be obtained with Equation (\ref{eq4}-\ref{eq8}) where \textit{W} is the deconvolution filter, (\textit{y}, \textit{x}) is the original filter coordinate and (\textit{y$_n$}, \textit{x$_n$}) is the new coordinate. 

\begin{equation}
N = s^2\label{eq3}
\end{equation}

\begin{equation}
n = s \times mod(y,s) + mod(x,s)\label{eq4}
\end{equation}

\begin{equation}
W_{n}[y_{n}][x_{n}] = W[y][x]\label{eq5}
\end{equation}

\begin{equation}\label{eq6}
\left\{
\begin{aligned}
x_{n} = K_{T} - ceil(x~/~s)\\
y_{n} = K_{T} - ceil(y~/~s)
\end{aligned}
\right.
\end{equation} 

\par\setlength\parindent{1em} where
\begin{equation}\label{eq7}	
\left\{
\begin{aligned}
0 \leq x < K + P_{K}\\
0 \leq y < K + P_{K}
\end{aligned}
\right.
\end{equation}
\begin{equation}\label{eq8}	
\left\{
\begin{aligned}
0 \leq x_{n} < K_{T}\\
0 \leq y_{n} < K_{T}
\end{aligned}
\right.
\end{equation}

\begin{center}
	$n \in \{0, 1, 2, ...N-1\}$
\end{center}

\par\setlength\parindent{1em}Step 1 and Step 2 basically split the deconvolution filters to multiple small convolution filters. This needs to be done only once and can be reused. Therefore, they can be done off-line with software approach. Unlike the first two steps, Step 3 and 4 are performed on the CNN processors for each input feature map. In step 3, the input feature maps also need to be padded with zeros to obtain equivalent deconvolution output. Otherwise, the output activations on the edge will be ignored. \textit{P$_I$} columns/rows of zeros will be added where \textit{P$_I$} is obtained from Equation (\ref{eq9}).

\begin{equation}
P_{I} = K_{T} - 1\label{eq9}
\end{equation}

\begin{figure}[t]
	\centerline{\includegraphics[height=4.8cm,width=8.8cm]{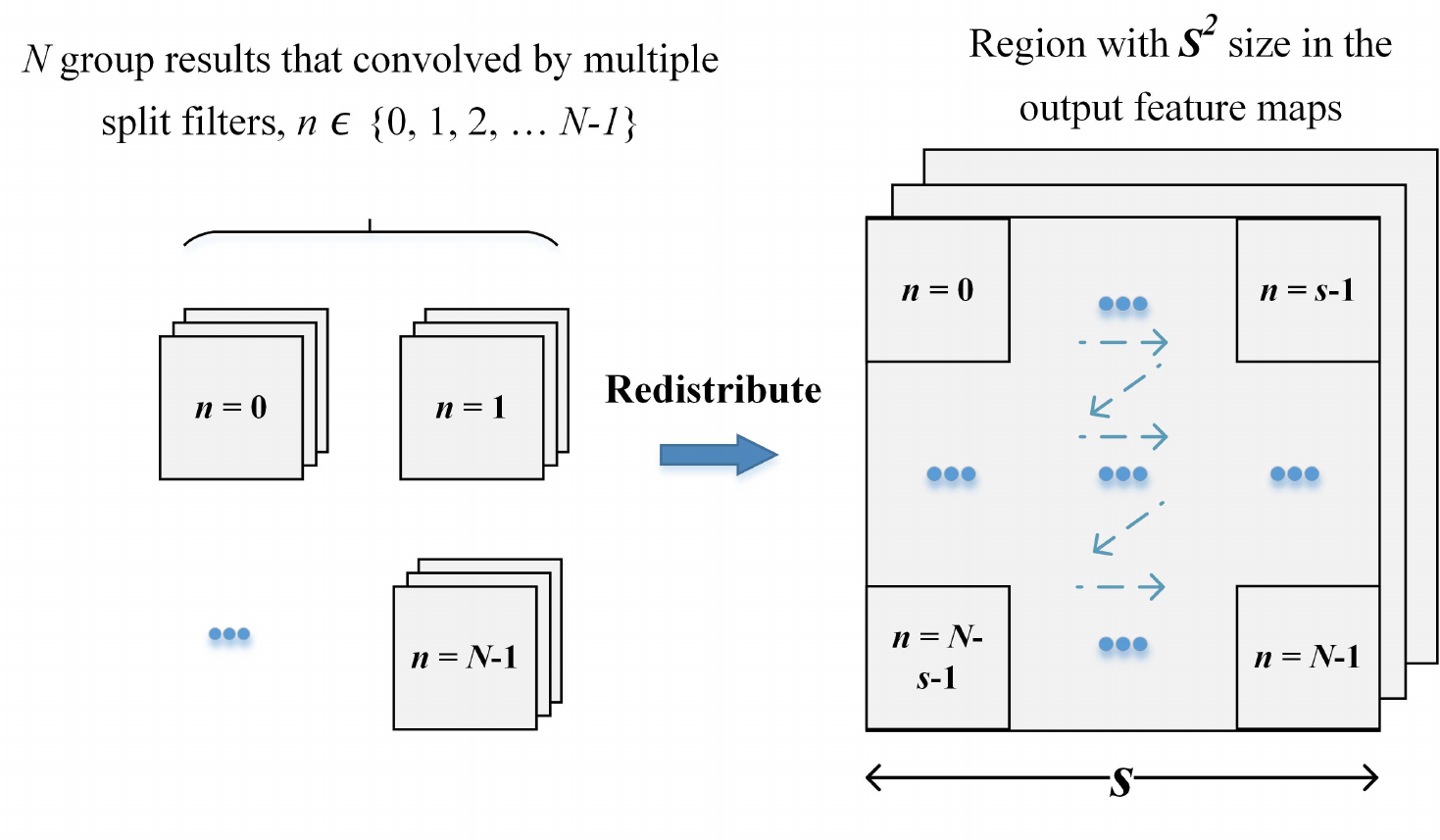}}
	\caption{Demonstrate the redistribution process for multiple groups of output activations.}
	\label{fig7}
\end{figure}

Finally, the N split convolution outputs need to be merged to form the deconvolution output. The reorganization pattern is illustrated in Figure \ref{fig7} and formulated in Equations (\ref{eq10}-\ref{eq13}). Contrary to the filter splitting process, we pick an element of each convolution output to construct an s×s window in the deconvolution output. Note that $ConvO$\underline{ }$n[x_i][y_i]$ represents the nth split convolution output and $O[x_f][y_f]$ refers to the expected deconvolution output. Suppose $(y_i, x_i)$ is coordinate of a split convolution output and $(y_f, x_f)$ is the coordinate of deconvolution output. The reorganization here does not need additional hardware as long as the partial convolution output can write the buffers with stride s which is usually allowed in generic CNN processors supporting tiling.

\begin{equation}
O[x_f] =ConvO\_n[x_i] \times s + mod(n~/~s)\label{eq10}
\end{equation}
\begin{equation}
O[y_f] =ConvO\_n[y_i] \times s + floor(n~/~s)\label{eq11}
\end{equation}
\par\setlength\parindent{1em} where
\begin{equation}\label{eq12}	
\left\{
\begin{aligned}
0 \leq x_i < I + 2P_I - K_T + 1\\
0 \leq y_i < I + 2P_I - K_T + 1
\end{aligned}
\right.
\end{equation}
\begin{equation}\label{eq13}	
\left\{
\begin{aligned}
0 \leq x_f < (I + 2P_I + 1) \times s + K + P_K\\
0 \leq y_f < (I + 2P_I + 1) \times s + K + P_K
\end{aligned}
\right.
\end{equation}
\par\setlength\parindent{1em}With the above four steps, we can convert generic deconvolution operations to split convolution operations and apply deconvolution on an unmodified CNN processor. In spite of the hardware compatibility, the proposed split deconvolution approach may extend the filters and input feature maps, which will induce additional computing overhead. On the other hand, the padding are zeros and can be potentially skipped by the conventional CNN processor optimizations. Detailed evaluation on realistic benchmarks will be discussed in the experiments.
\section{Experiments}
\par\setlength\parindent{1em}This section consists of three parts. First, we listed the setting of the selected benchmarks and the experimental environment. Then we evaluate the performance and energy consumption of split deconvolution on the generic propose processors. At last, the approach is compared with two off-the-shelf processors i.e. Google Edge TPU and Intel NCS2. The proposed SD algorithm and its deployment on the neural network processors are open sourced and can be found in https://github.com/warmthless/split-deconvolution.

\subsection{Experimental setup}
\par\setlength\parindent{1em}To perform comprehensive evaluation of the proposed split deconvolution computing approach, we conduct experiments on both simulation-based neural network processors and commodity neural network processors provided by the chip vendors, and then compare proposed methods with prior deconvolution computing approaches.
\par\setlength\parindent{1em}For the simulation-based evaluation, we developed cycle-accurate neural network simulators for both the dot-production based neural network processor architecture and the regular 2D array architecture. Both the 8-bit dot-production PE array and the 2D PE array are implemented and synthesized with Synopsys Design Compiler (DC) under TSMC 40nm library. The dot-production based architecture includes 16 processing units, and each unit performs dot production on 16 input activations and weights. The 2D PE array is set to be 32 by 7. The I/O buffer size is set to be 256 KB, weight buffer is 416 KB. Both processors run at 800 MHz.
\par\setlength\parindent{1em}For the commodity neural network processors, we choose two representative ones. One of them is Edge TPU \cite{jouppi2017datacenter} from Google and it does not support native deconvolution operations. To implement deconvolution on it, we convert the deconvolution to standard convolution using zero padding [6]. The other processor chip is the latest NCS2 \cite{ncs2} from Intel. It supports native deconvolution operation and the deconvolution is applied directly on the optimized architecture of NCS. The performance on the commodity processors is measured using the system clock.
\begin{table}[b]
	\caption{  Comparison of multiply-add operands (deconvolution layers) for three different implementations}
	\centerline{\includegraphics[height=3.2cm,width=8.8cm]{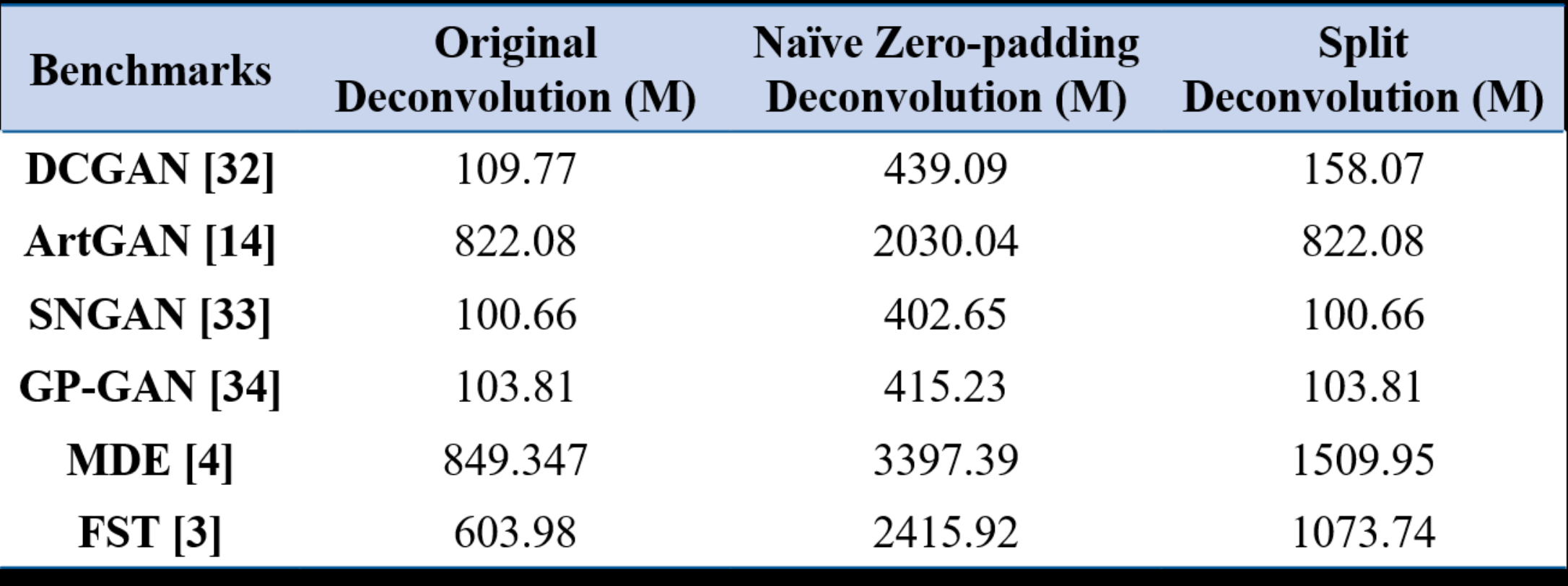}}
	\label{tab2}
\end{table}

\begin{table}[t]
	\caption{  Comparison of weight parameters (deconvolution layers) for three different implementations }
	\centerline{\includegraphics[height=3.2cm,width=8.8cm]{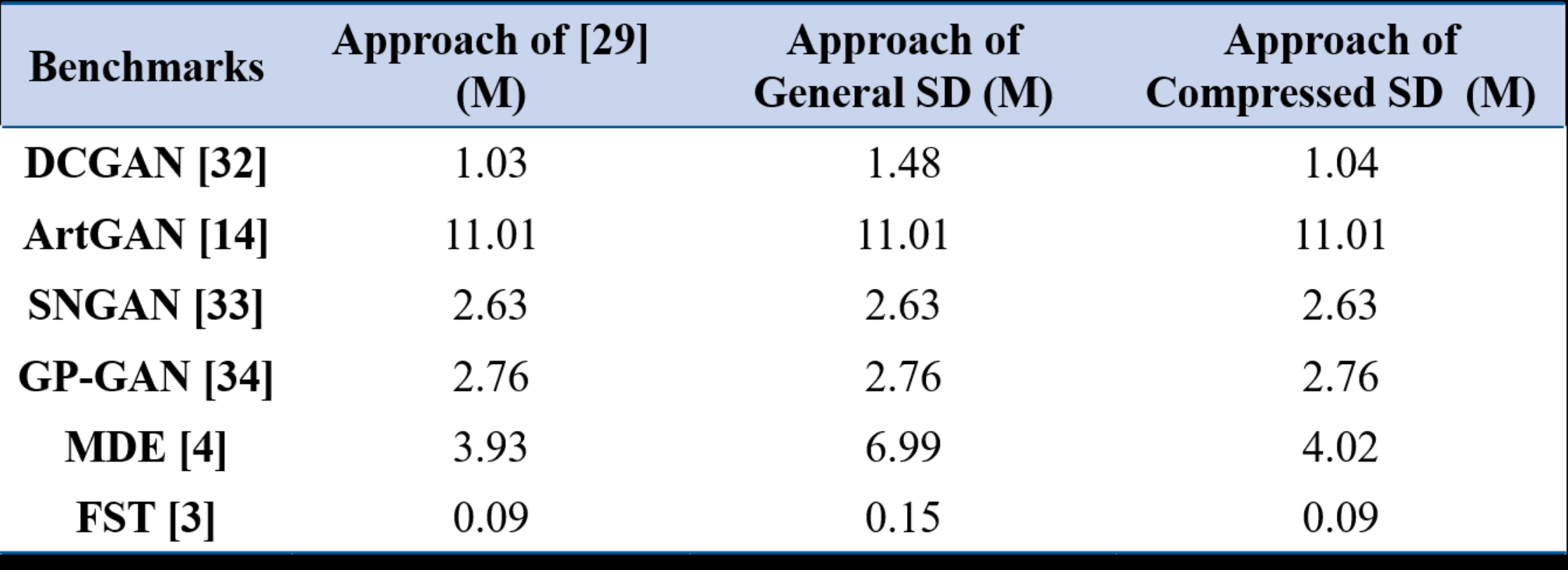}}
	\label{tab8}
\end{table}

\par\setlength\parindent{1em}To evaluate the different deconvolution approaches, we selected a set of advanced neural networks as our benchmarks including ArtGAN \cite{tan2017artgan} on Cifar 10 (ArtGAN), DCGAN \cite{radford2015unsupervised} on Large-scale CelebFaces Attributes Dataset (DCGAN), Spectral Normalization for GAN \cite{miyato2018spectral} on Cifar 10 (SNGAN), GP-GAN on Transient Attributes Database \cite{wu2017gp} (GP-GAN) for generating new datasets. Unsupervised Monocular Depth Estimation of FCN on KITTI and Cityscapes \cite{godard2017unsupervised} (MDE) aims of image segmentation, and Fast-Style-Transfer \cite{lengstrom2016faststyletransfer} on CoCo2014 which is used to apply the style  of one image to another image (FST).   

\subsection{Experimental results on general CNN processors}
\par\setlength\parindent{1em}This section illustrates how the proposed split deconvolution improves the performance and efficiency of generative neural networks on the simulated general CNN processors including both dot-production array and 2D array architectures.

\subsubsection{\textbf{Operation number and parameters comparison}}
\par\setlength\parindent{1em}Multiply-add (MAC) operation takes up the majority of the computing in neural networks, so the number of MACs exhibits the computing intensity of the neural networks directly and it is independent with the underlying computing architectures. Thereby, we use this metric to compare the different deconvolution computing approaches. Table \ref{tab2} shows the number of MACs in original neural networks, neural networks using native zero padding (NZP) and neural networks using the proposed split deconvolution (SD). It can be observed that NZP incurs a large number of redundant operations compared with the original deconvolution. Compared to NZP, SD brings in much less computing. It does not incur any additional computing overhead in SNGAN, ArtGAN and GP-GAN and induces only a portion of additional computing on the rest of the neural networks. 
\par\setlength\parindent{1em}In theory, SD will not increase the amount of the computation. The deformation approach proposed in \cite{8806783} transforms filter into different shapes, which does not introduce redundant parameters. But for some of the legacy accelerators, they may not support filters with irregularly shapes in the layer. On this occasion, zeros need to be added to further apply to map deconvolutional layers on general-propose architectures. When the original filter length or width is not divisible by the stride s in the according neural networks, we need to pad zeros on the top and left side of the filters to ensure identical filter splitting. This neat zero value can be easily compressed on an accelerator with a particular data format. Table \ref{tab8} lists the number of weight parameters of original neural networks \cite{8806783}, general SD approach and SD with compressed weight parameters. Though there are induced zeros of weight in some benchmarks (DCGAN, MDE and FST), most of the redundant values have been been removed after the compression. In addition, the split deconvolution may produce only the 
center area of the original deconvolution output feature maps, and we must add zero paddings to the input feature maps to obtain equivalent deconvolution output feature maps. Thereby, the proposed split deconvolution may add zeros to both the weights and the input activations, and induce more computing depending on the neural network parameters. 
\par\setlength\parindent{1em}The purpose of padding zeros on filters and input feature maps is to make the SD approach to be a more general solution to accelerate GANs on legacy accelerators. Meanwhile, this induced redundant values can be omitted and have no impact on performance, which is analyzed in detail in the next section. 

\subsubsection{\textbf{Performance comparison}}
\begin{figure}[t]
	\centerline{\includegraphics[height=3.8cm,width=8.8cm]{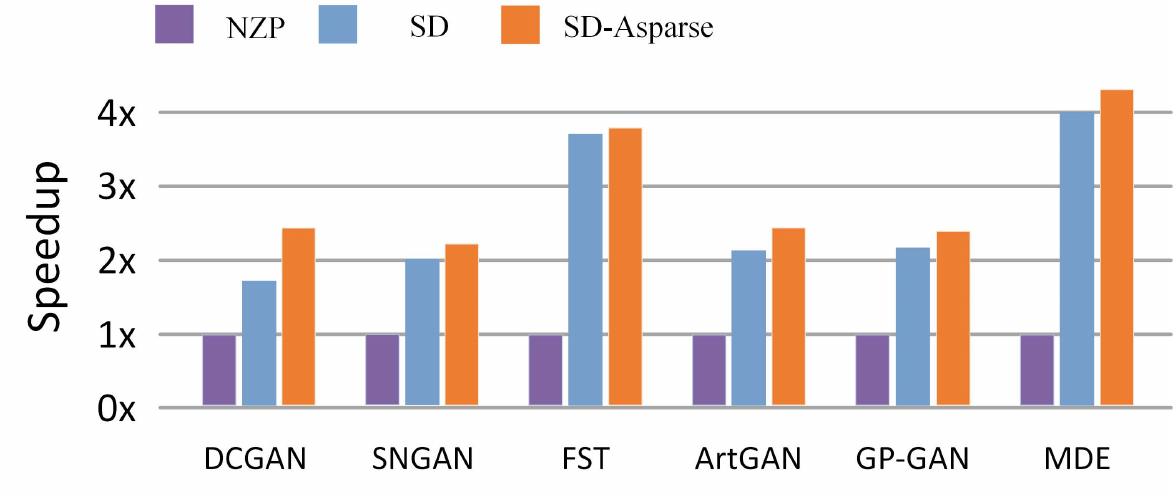}}
	\caption{Performance comparison of the deconvolutional layers in the dot-production PE array.}
	\label{fig8}
\end{figure}
\begin{figure}[t]
	\centerline{\includegraphics[height=3.6cm,width=8.8cm]{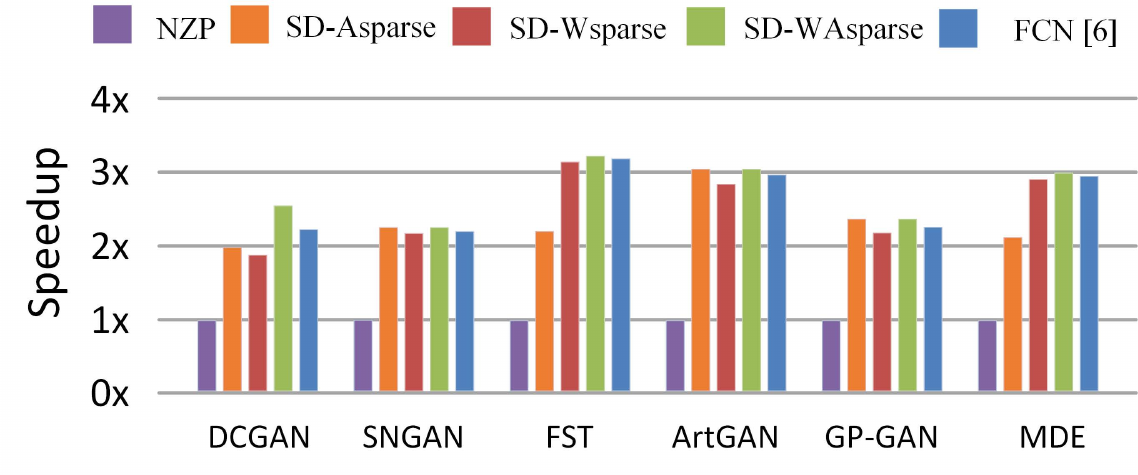}}
	\caption{ Performance comparison of the deconvolutional layers in the regular 2D PE array.}
	\label{fig9}
\end{figure}
\par\setlength\parindent{1em}In this section, we mainly compare the different deconvolution approaches on typical neural network processors. Although NZP and SD may induce redundant computing, many of the redundant computing can be potentially squeezed using the sparse aware optimization techniques which allow the processors to skip the zero multiplications. Generally, there are three different sparse-aware optimization methods including activation sparse optimization (Asparse), weight sparse optimization (Wsparse) and activation and weight sparse optimization (AWsparse). We explored the neural network performance on processors with the different optimization methods. While the processor with dot-production PE array cannot skip zero weights and we only apply Asparse method on it. In addition, we also compare with  FCN-engine 
\cite{xu2018fcn} that had the 2D PE array CNN processor redesigned. 
\par\setlength\parindent{1em}Figure \ref{fig8} depicts the normalized performance of three acceleration schemes on the dot-production PE array. NZP incurs 75\% computing redundancy on average on the benchmark neural networks when converting the deconvolution to convolution. Unlike the NZP, split deconvolution has only marginal zero paddings on the boundary in some corner cases. Therefore, it has much less computing redundancy, which is projected in the 2.5$\times$ performance boost of SD over NZP. When the specified input activation lines can be skipped to generate standard deconvolution output, the performance can further be improved. Notably, SD-Asparse on DCGAN improves by 1.4$\times$. The primary reason lies in the fact that the DCGAN has fewer network layers and smaller input feature maps. As a result, the computing redundancy caused by the padding affects the overall performance more significantly. 
\par\setlength\parindent{1em}On the 2D PE array CNN processor as shown in Figure \ref{fig9}, SD-Asparse and SD-Wsparse in the experiments show the influence of the filter expansion and the input expansion respectively. Although SD-Wsparse induces some redundant computation due to padding to the input feature maps, most of the convolution processors support zero-skipping and can squeeze the computing redundancy automatically. Compared to SD-Wsparse, SD-WAsparse that enables the zero-skipping reduces 22\% redundant computation on average. Similarly, SD-Asparse has zero-padding added to the weights, and the redundant computing can also be eliminated on a sparse convolution processor architecture. For workloads like DCGAN, FST, and MDE, the filters need to be expanded. In these cases, SD-WAsparse reduces 75\% - 80\% computing redundancy with zero-skipping. When the split deconvolution is deployed on optimized CNN processors, the performance of SD-WAsparse is on par with that of FCN in all the benchmark neural networks. The deconvolution approach presented in FCN-engine [6] adopts a bi-directional data flow. It has implemented the original deconvolution, which is the input activations multiplied with each filter and then accumulates the overlapped production. By taking advantage of the column buffers, it can transmit the partial results for accumulation efficiently. However, the output feature maps on edge are redundant and need to be cropped, which inevitably induces computing overhead, especially for smaller deconvolution layers. Therefore, SD-WAsparse outperforms FCN-engine on some of the neural networks like DCGAN, as shown in Figure \ref{fig9}.

\subsubsection{\textbf{Energy consumption comparison}}
\begin{figure}[t]
	\centerline{\includegraphics[height=4.2cm,width=8.8cm]{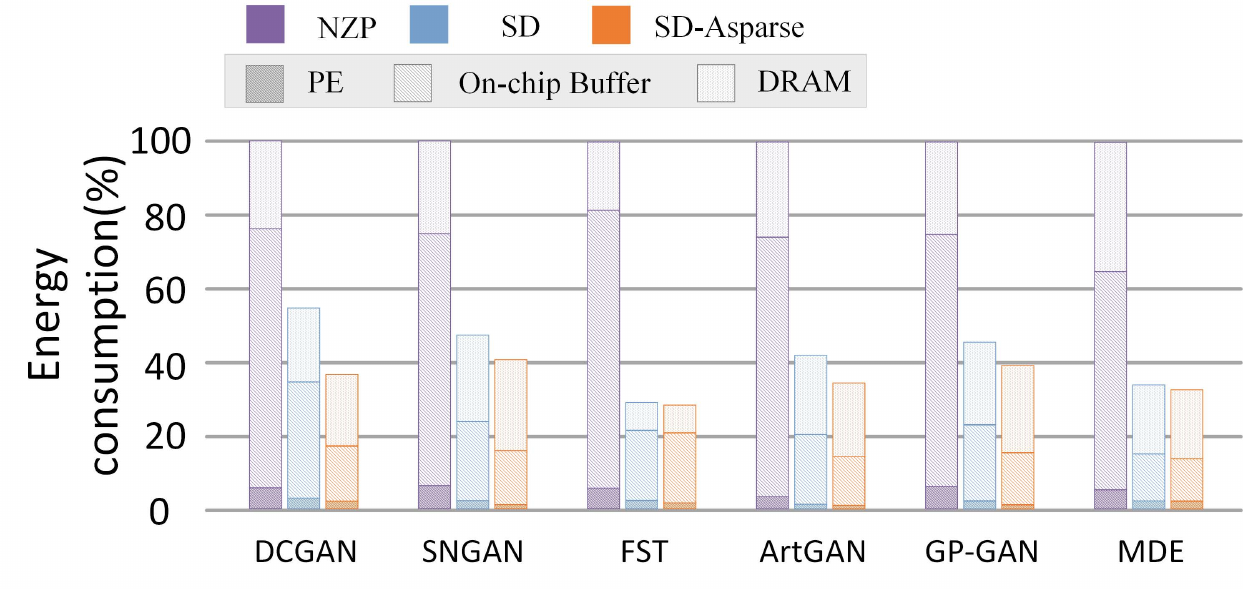}}
	\caption{Energy consumption of the deconvolutional layers in the dot-production PE array.}
	\label{fig10}
\end{figure}

\begin{figure}[t]
	\centerline{\includegraphics[height=4.4cm,width=8.8cm]{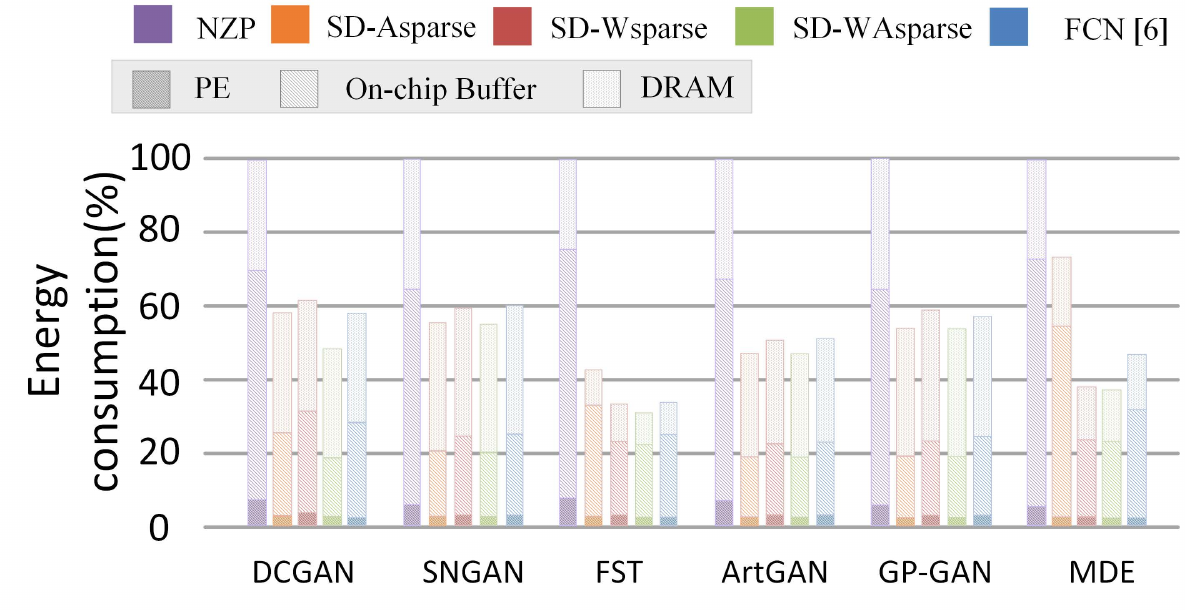}}
	\caption{Energy consumption of the deconvolutional layers in the regular 2D PE array.}
	\label{fig11}
\end{figure}
\par\setlength\parindent{1em}Figures \ref{fig10} and \ref{fig11} present the relative energy consumption distribution of the different deconvolution approaches on the dot-production PE array and regular 2D PE array respectively. Compared to NZP, the average energy consumption of SD-Asparse and SD-WAsparse reduce by 36.15\% and 43.63\% respectively on the two CNN architectures. Unlike the performance comparison, the energy consumption comparison is less significant. In general, the deconvolution energy consumption roughly consists of three parts i.e. PE, on-chip buffer, and DRAM. According to the estimation using CACTI \cite{li2011cacti}, the energy is mostly consumed by the DRAM access and the on-chip buffer access. While the amount of DRAM access of the different deconvolution approaches is about the same, their consumption has little difference across these approaches. Despite the dramatic difference in PE activity and energy consumption, PE energy consumption is too small to affect the overall deconvolution energy consumption. As a result, the energy consumption difference is primarily determined by the amount of on-chip buffer accesses, which explains all the energy consumption difference. For example, SD-Asparse induces relatively more weight reading and thus higher energy consumption. Similarly, FCN requires additional on-chip buffers to support the unified convolution and deconvolution, so the overall energy consumption is higher than that of SD-WAsparse in all the benchmark networks, though their performance is quite close to each other.

\subsubsection{\textbf{SD-based DCGAN Demo}}
\par
\begin{figure}[t]
	\centerline{\includegraphics[height=6.6cm,width=8.8cm]{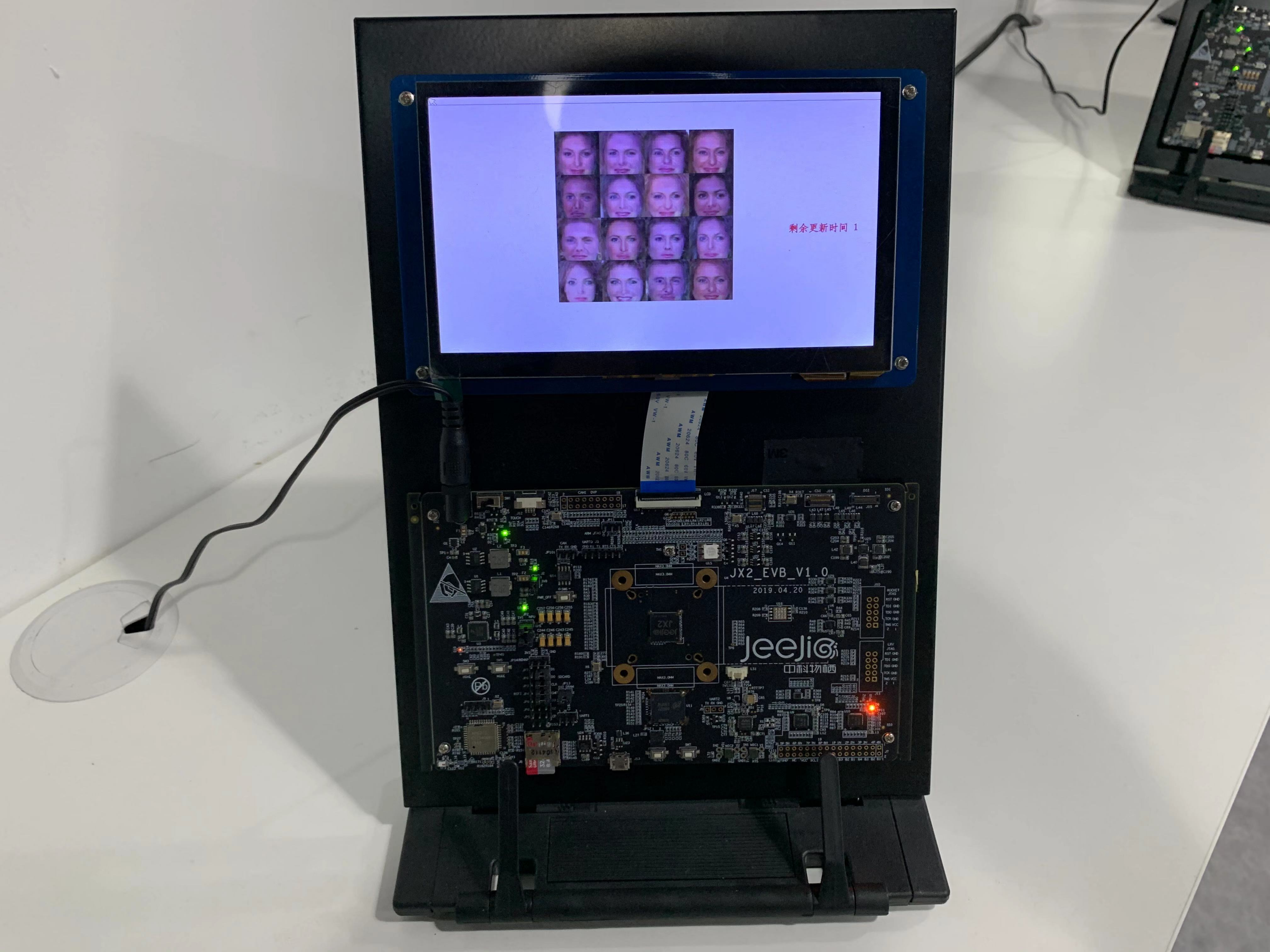}}
	\caption{The DCGAN demo on edge system jeejio JX2\cite{jeejio}}
	\label{jeejio}
\end{figure}

\setlength\parindent{1em} As we need to manipulate the output write instruction of the neural network processor to reorganize the split deconvolution outputs for the equivalent deconvolution output, we apply the proposed SD approach on an edge AI system of which we can touch the low-level output write instructions to demonstrate the use of the proposed SD algorithm. Note that the edge AI is produced by \cite{jeejio}. It consists of both RISC-V cores and a neural network processor fabricated with TSMC 40nm technology. Each split convolution is executed sequentially on the neural network processor and the conventional sequential output write instruction is replaced with a stride write instruction which is widely supported in DMA cores. On this AI system, we implemented a face generation demo using DCGAN as shown in Figure \ref{jeejio}. The end-to-end performance comparison with NZP is consistent with that obtained in Figure \ref{fig9}, which demonstrates the computing efficiency of SD on neural network processors.

\subsubsection{\textbf{Deconvolution Conversion Quality Evaluation}}
\par
\begin{figure}[h]
	\centerline{\includegraphics[height=2.35cm,width=6.8cm]{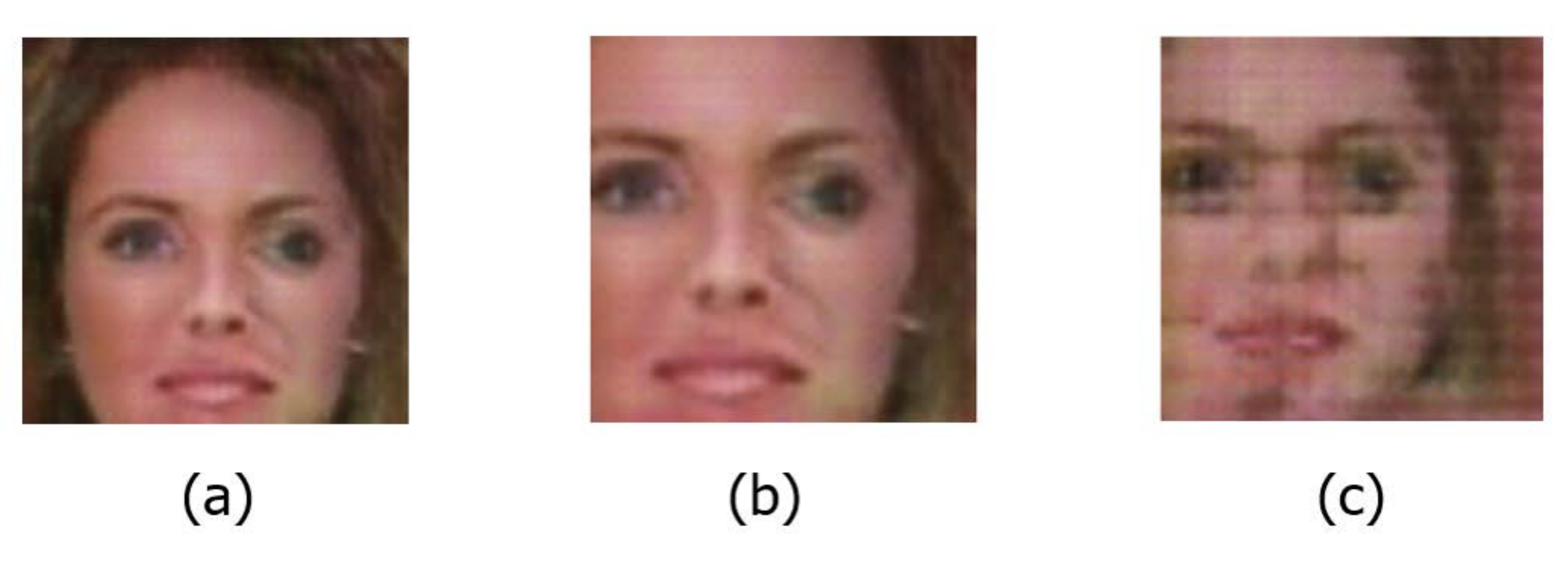}}
	\caption{The generated images of DCGAN\cite{radford2015unsupervised}. (a) SD approach (b) The approach base on \cite{shi2016deconvolution} (c) The approach based on \cite{chang2018optimizing}}
	\label{dcgan}
\end{figure}

\begin{figure}[h]
	\centerline{\includegraphics[height=2.35cm,width=6.8cm]{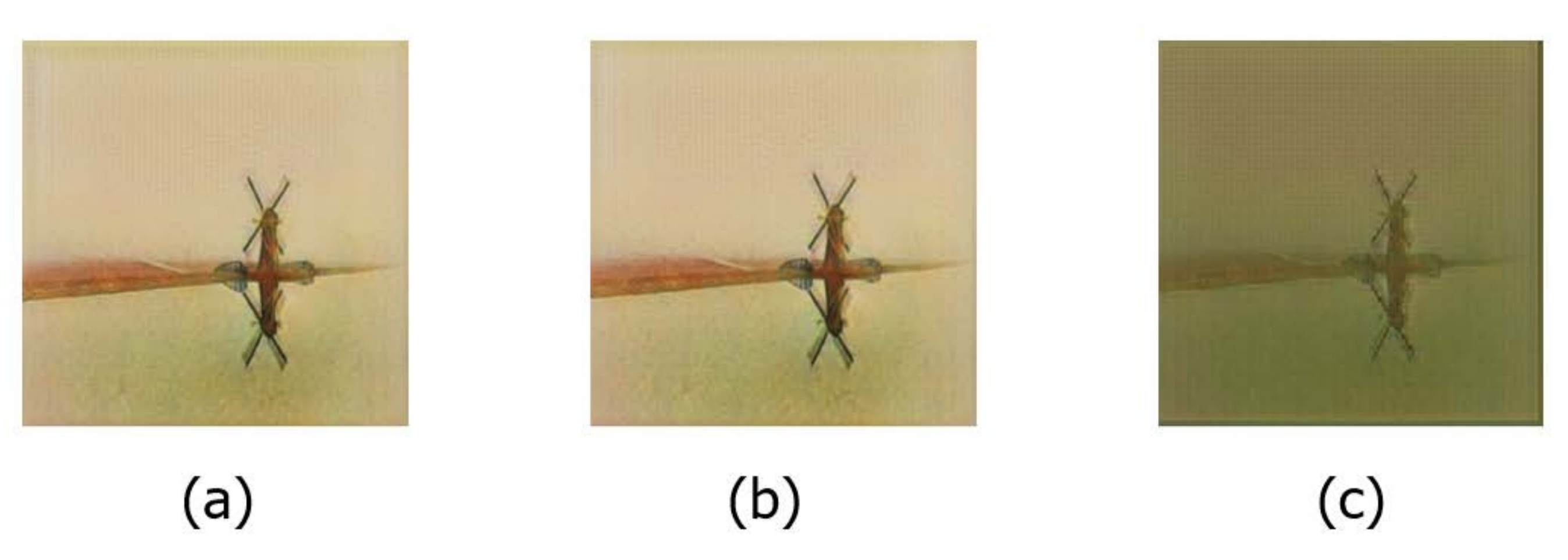}}
	\caption{The generated images of FST\cite{lengstrom2016faststyletransfer}. (a) SD approach (b) The approach base on \cite{shi2016deconvolution} (c) The approach based on \cite{chang2018optimizing}}
	\label{fst}
\end{figure}

\setlength\parindent{1em} In this section, we mainly evaluate the quality of the deconvolution results calculated using different deconvolution conversion approaches. We compare the results to that generated from the raw deconvolution with SSIM metric \cite{wang2004image} which is widely utilized to measure the similarity between images. SSIM ranges from 0 to 1 and higher SSIM indicates higher similarity between the images. The comparison is shown in Table \ref{tab3}. It can be observed that SD produces identical results for both DCGAN and FST, while the methods proposed in \cite{shi2016deconvolution} and \cite{chang2018optimizing} produce different results and the SSIM of the same deconvolution conversion approach varies on different generative neural network models. Particularly, the approach in \cite{shi2016deconvolution} results in considerable computing errors in DCGAN while minor computing errors in FST. This is mainly caused by the fact that the input images in FST are larger and the influence of the wrong padding on the boundaries is less significant. Moreover, the proportion of deconvolution in FST is smaller than that in DCGAN, which also explains the higher SSIM metric in FST.
\begin{table}[b]
	\caption{SSIM value comparison}
	\centerline{\includegraphics[height=1.5cm,width=8cm]{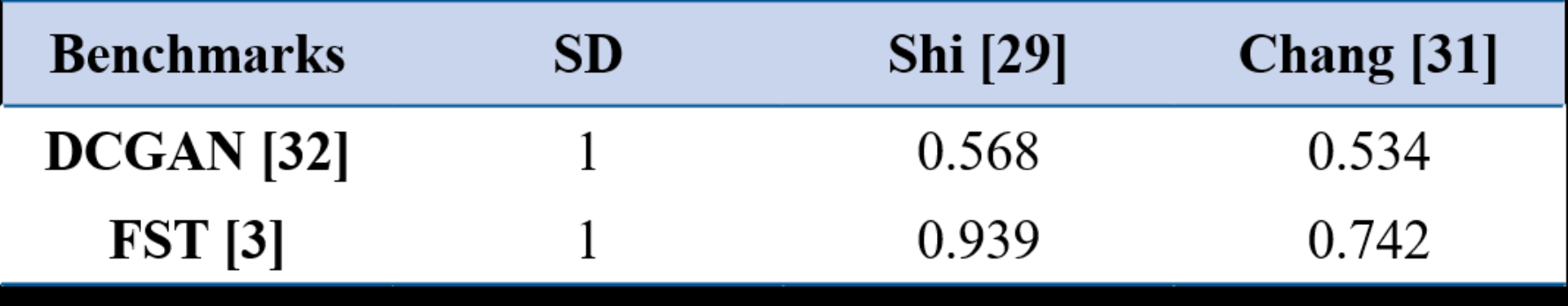}}
	\label{tab3}
\end{table}
To further illustrate the effect of the errors on the generated images, we also display the images generated using different deconvolution approaches in Figure \ref{dcgan} and \ref{fst}. It can be seen that the quality of the images calculated with the different deconvolution conversion approaches is roughly consistent with the SSIM metric. Basically, FST using the deconvolution conversion approach proposed in \cite{shi2016deconvolution} seems to be acceptable visually. However, the generated images in the rest cases differ dramatically and can not be tolerated or utilized. 


\subsection{Experiments on commodity NN processors}
\par\setlength\parindent{1em}This section shows the use of SD for generative neural networks on the most advanced commodity CNN processor chips including Google Edge TPU without specialized deconvolution support and Intel NCS2 with specialized deconvolution operation support. As we cannot touch the internal output write instructions in these commodity neural network processors, we can not reorganize the split convolution results for the equivalent deconvolution outputs directly though the stride output write is probably supported. To demonstrate the use of SD on these neural network processors, we move the results generated in each split convolution to host and have the host to reorganize the results for the following neural network operations. Since the data movement is not required given internal data movement support in the neural network processors, we only take the split deconvolution computing time and the data reorganization time as the overall deconvolution execution time in the experiments.

\begin{figure}[t]
	\centerline{\includegraphics[height=4.2cm,width=8.8cm]{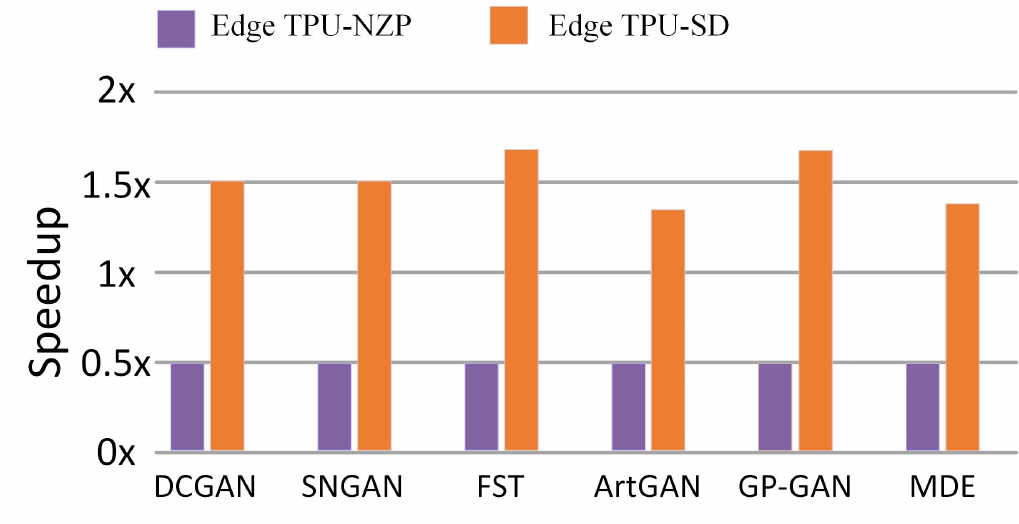}}
	\caption{ Performance comparison on Edge TPU.}
	\label{fig15}
\end{figure}

\begin{figure}[t]
	\centerline{\includegraphics[height=4.2cm,width=8.8cm]{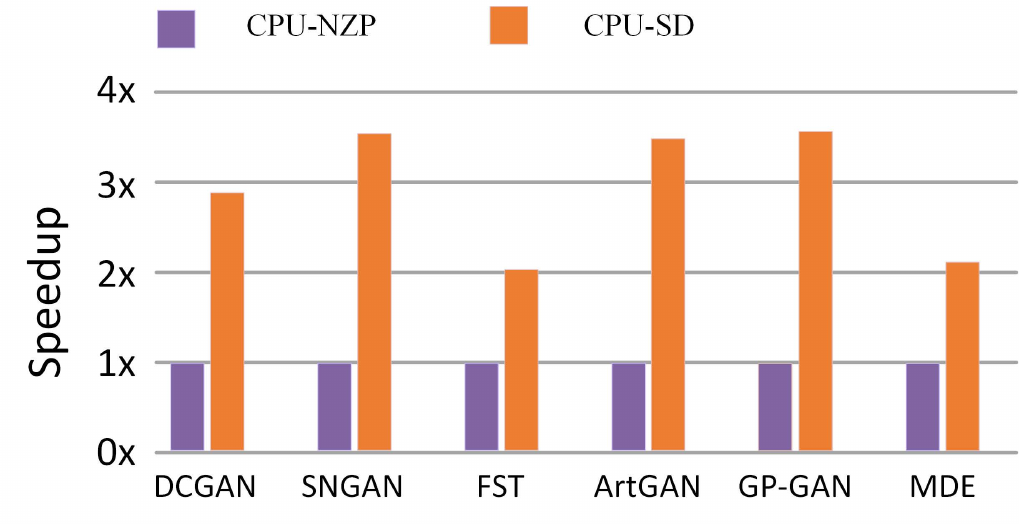}}
	\caption{ Performance comparison under the same computational efficiency on host.}
	\label{fig16}
\end{figure}

\subsubsection{\textbf{Edge TPU}}
\par\setlength\parindent{1em}Edge TPU is a tensor processor with the systolic array architecture and is usually used as a co-processor of a host computer. It does not support native deconvolution operation, so we apply the NZP approach to implement the deconvolution on it as the baseline. Meanwhile, we also perform split deconvolution (SD) to deploy deconvolution on TPU. The NZP approach requires zero-padding to the input feature maps, and the SD approach needs additional output feature reorganization. While this computing cannot be performed on TPU directly, we have them done on the host processor. 

\par\setlength\parindent{1em}The normalized acceleration performance of the two deconvolution approaches on Edge TPU is illustrated in Figure 
\ref{fig15}. The proposed SD achieves 1.51$\times$ performance speedup over NZP on average. Particularly, FST yields the highest speedup (1.65$\times$) over NZP on Edge TPU. However, the performance improvement is much lower than that on CPU, which is not consistent with the number of operations as listed in Table \ref{tab2}. To explore the underlying reasons, we further evaluate the computing efficiency of convolution with different input feature map sizes and filter sizes on Edge TPU. The evaluation result is revealed in Table \ref{tab4} and Table \ref{tab5}. The filter size is set to be 3$\times$3, which is a frequent setup for split deconvolution and measure the Giga multiply-add operations per second (GMACPS) given different input feature maps. As shown in Table \ref{tab4}, when the size of input feature maps ranges from 8$\times$8 to 128$\times$128, the normalized computational efficiency of Edge TPU i.e. GMACPS increases significantly. Although there are not much documents about the detailed computing architecture of Edge TPU, it is probably that Edge TPU compiler parallelizes the convolution operations on the 2-D plane of the input features and requires larger input feature maps to make good use of its computing resources. Similarly, we also investigate the influence of filter sizes on the computing efficiency in Table \ref{tab5}. We set the feature map to be 128$\times$128 and change the filter size from 2$\times$2 to 5$\times$5. When we compare the computing efficiency, it can be observed that the convolution with larger filter sizes on Edge TPU is clearly more efficient. For SD that splits the deconvolution operations to multiple smaller convolution operations, the resulting convolution is usually less efficient compared to the NZP based converter. Basically, the computing efficiency of Edge TPU degrades with smaller feature maps and filter sizes due to its inherent convolution parallelization approach. Thereby, the performance speedup of SD over NZP is lower than that is estimated based on the number of MACs.  

\par\setlength\parindent{1em}To further verify the above analysis, we have both models of NZP and SD run on the host CPU of which the computing efficiency does not vary much under different kernel parameters. The normalized acceleration performance of the NZP and SD approaches is illustrated in Figure \ref{fig16}. (Note that the host processor is Intel Core i7-7700 with 3.6 GHz.) It can be found that the proposed SD achieves 3.04$\times$ performance speedup over NZP on average, which is roughly consistent with the magnitude of operation reduction presented in Table \ref{tab2}. Particularly, the performance speedup goes up to 3.60$\times$ on GP-GAN. Similar to Figure \ref{fig9}, the average performance improvement of DCGAN, FST, and MDE is relatively lower than that of SNGAN, ArtGAN, and GPGAN due to the additional parameters padded to the filters and the input features during splitting. This confirms the analysis that SD does reduce the amount of computing compared to that in NZP but the converted convolution with smaller kernel sizes and lower computing efficiency affects the performance speedup. If the neural network processors improve its computing efficiency for smaller convolution kernel sizes, the performance speedup of SD over NZP will be higher accordingly.

\begin{table}[t]
	\caption{Normalized GMACPS for different input feature map size on Edge TPU}
	\centerline{\includegraphics[height=2.5cm,width=8.8cm]{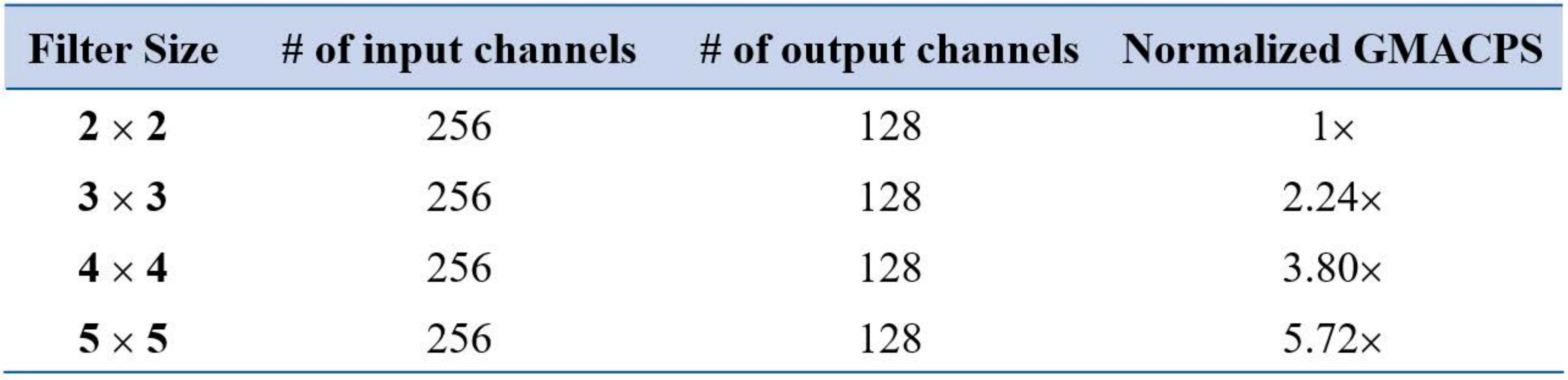}}
	\label{tab4}
\end{table}
\begin{table}[t]
	\caption{Normalized GMACPS for different filter size on Edge TPU}
	\centerline{\includegraphics[height=2.3cm,width=8.8cm]{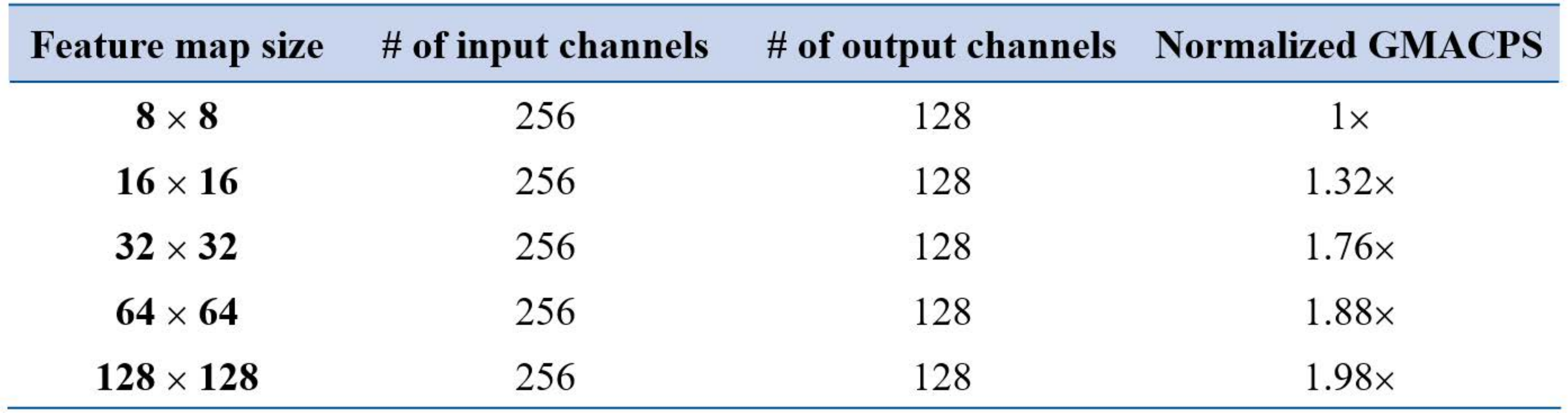}}
	\label{tab5}
\end{table}

\subsubsection{\textbf{Intel Neural Compute Stick 2 (NCS2)}}
\par
\begin{figure}[t]
	\centerline{\includegraphics[height=4.2cm,width=8.8cm]{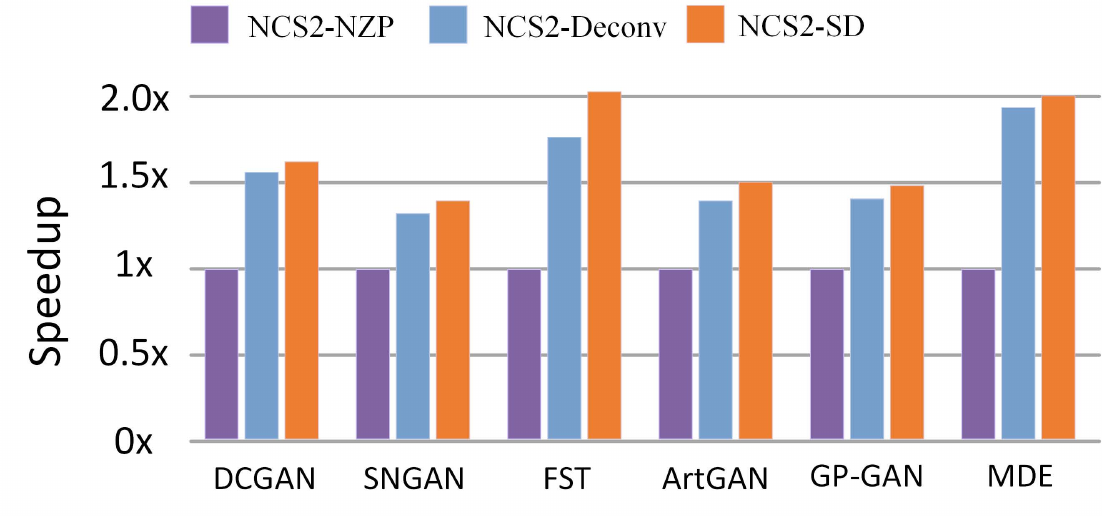}}
	\caption{Performance comparison on Intel Neural Compute Stick 2.}
	\label{fig17}
\end{figure}

\setlength\parindent{1em}NCS2 is a neural network processor produced by Intel, and it includes specialized hardware to support native deconvolution operation. We evaluated the deconvolutional layers of generative neural networks on it with the deconvolution operations implemented using the NZP approach, the SD approach as well as the native deconvolution. The experiment is presented in Figure \ref{fig17}. When compared to NZP, the proposed SD performs 1.67$\times$ performance speedup over the NZP approach. Similarly to Edge TPU, its performance speedup is lower than that analyzed with MACs. Therefore, we also evaluate the influence of different feature map size and filter size and the result is shown in Table \ref{tab6} and Table \ref{tab7} with the same configurations as Edge TPU. And we notice that the lower computing efficiency of smaller convolution kernel sizes on NCS2 is the major reasons for the lower performance speedup. 

\par\setlength\parindent{1em}While NCS2 also includes specialized hardware for native deconvolution operation, we further evaluated the deconvolutional layers of generative neural networks on it with the optimized deconvolution. Even compared to the native deconvolution implementation on NCS2, the proposed SD approach still yields 1.10$\times$ performance speedup on average. Despite the degraded computing efficiency of NCS2 on the split convolution kernels, the proposed SD approach still show higher performance NCS2 without any hardware modification.
 
\begin{table}[b]
	\caption{Normalized GMACPS for different input feature map size on NCS2}
	\centerline{\includegraphics[height=2.5cm,width=8.8cm]{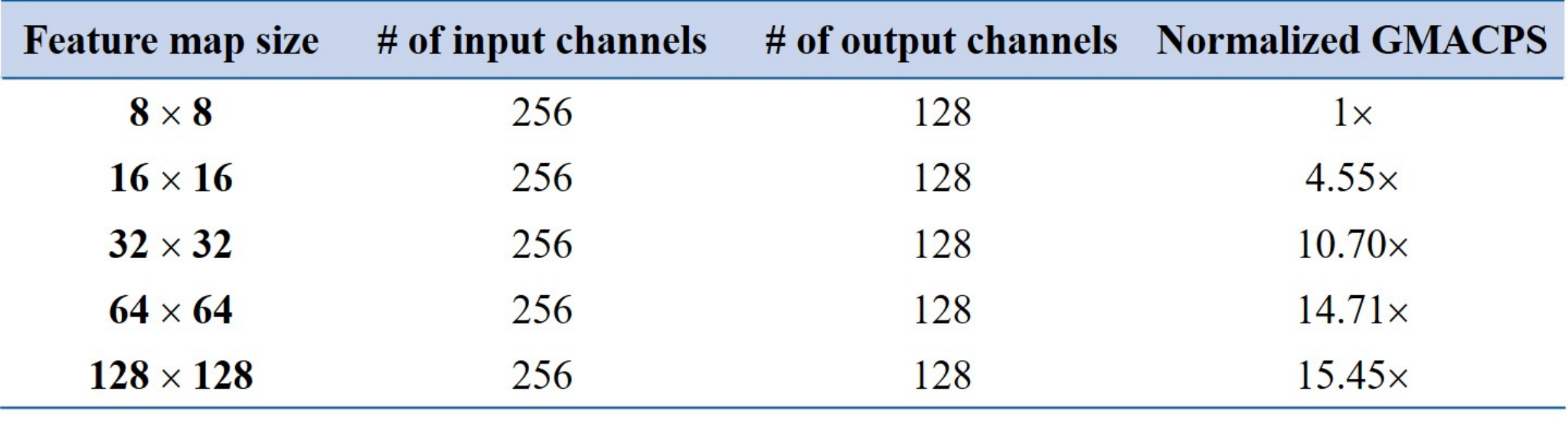}}
	\label{tab6}
\end{table}

\begin{table}[b]
	\caption{Normalized GMACPS for different filter size on NCS2}
	\centerline{\includegraphics[height=2.3cm,width=8.8cm]{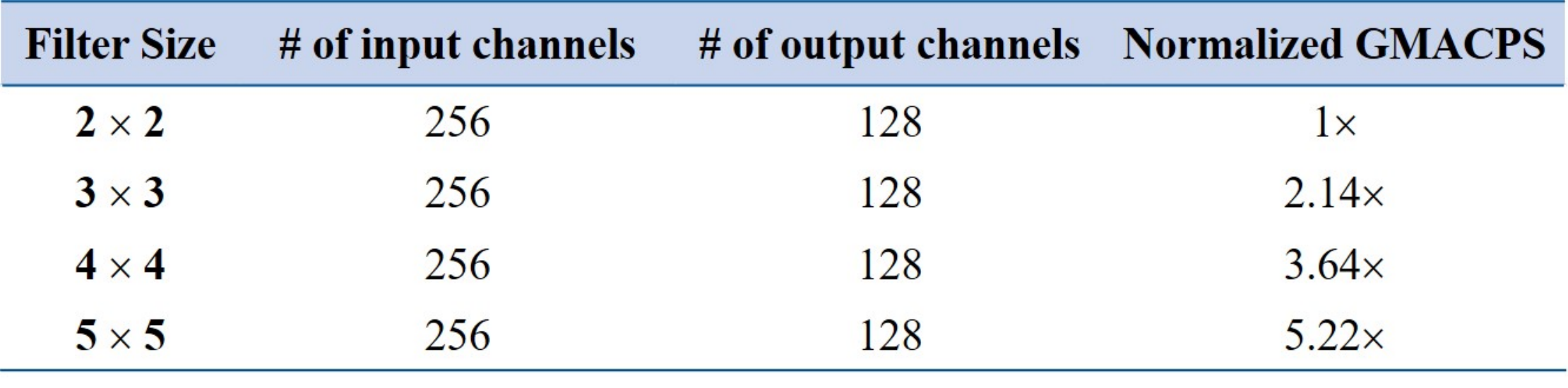}}
	\label{tab7}
\end{table}

\section{Conclusion}
\par\setlength\parindent{1em}Prior generative neural network acceleration may either require intensive hardware modification of existing CNN processors or bring in large amount of redundant computing because the involved deconvolution operations cannot be fitted to the conventional CNN processors directly. To address this problem, we proposes to convert the deconvolution to standard convolution with a software approach. The basic idea is to investigate the computing patterns of deconvolution and formulate it as convolution computing patterns. The resulting convolution filters can be obtained by splitting the original deconvolutional filters while the convolution results need to be reorganized to construct the original deconvolution results. This approach incur little computing redundancy, and thus enables fast and efficient deconvolution execution on legacy deep learning processors. With comprehensive experiments, we demonstrate that SD achieves 2.41$\times$ – 4.34$\times$ performance speedup over the naïve zero padding methods and is on par with the prior optimized implementation on modified fully convolution neural network processor. Moreover, the proposed approach is also beneficial to commodity neural processors. It yields 1.51$\times$ performance speedup compared to the naïve zero padding on Google Edge TPU which does not have native deconvolution support. When compared to Intel NCS2 chips with native deconvolution support, it still achieves 1.1$\times$ performance speedup on average though the computing efficiency of NCS2 degrades with the split convolution kernels.


\ifCLASSOPTIONcaptionsoff
  \newpage
\fi

%
\bibliographystyle{IEEEtran}
\bibliography{TC_Deconv}




\end{document}